\documentclass{article}
\usepackage{booktabs}
\usepackage{caption}
\usepackage{authblk}
\usepackage{amsmath}
\usepackage{amssymb}
\usepackage{amsthm}
\usepackage{graphicx}
\usepackage{multirow}
\usepackage[colorlinks=true, allcolors=blue]{hyperref}
\usepackage[margin=2cm]{geometry}
\usepackage[section]{placeins}
\usepackage[utf8]{inputenc}
\usepackage[T1]{fontenc}
\usepackage{lmodern}
\usepackage{microtype}

\title{Neural Latent Gaussian Process for 3D reconstruction of \\ Mouse Brain Lipids}

\author[1]{Daniel Trejo Banos $^{*,\dagger}$}
\author[2]{Artiom Casapu $^{*,\dagger}$}
\author[3]{Luca Fusar Bassini} 
\author[3]{Halima Hannah Schede} 
\author[3]{Giovanni D’Angelo}
\author[3]{Gioele La Manno}
\author[1]{Ekaterina Krymova $^{*}$}

\affil[1]{Swiss Data Science Center, Switzerland}
\affil[2]{ETH Zurich, Switzerland}
\affil[3]{EPFL, Switzerland}

\date{August 2026}

\begin{document}
\maketitle

\begingroup
\renewcommand{\thefootnote}{\fnsymbol{footnote}}
\footnotetext[1]{Corresponding authors:
\href{mailto:acasapu@student.ethz.ch}{acasapu@student.ethz.ch},
\href{mailto:daniel.trejobanos@sdsc.ethz.ch}{daniel.trejobanos@sdsc.ethz.ch}, 
\href{mailto:ekaterina.krymova@sdsc.ethz.ch}{ekaterina.krymova@sdsc.ethz.ch}.}
\footnotetext[2]{These authors contributed equally to this work.}
\endgroup

\section*{Abstract}

We address the problem of reconstructing continuous three-dimensional lipid profiles of the mouse brain from lipid measurements acquired only on a sparse set of two-dimensional tissue sections. Using MALDI-MSI measurements from coronal sections of two densely sampled and six sparsely sampled mouse brains, registered to the Allen Mouse Brain Common Coordinate Framework (CCF), we seek to interpolate the measured hundreds of lipid profiles across the unobserved voxels of the reference brain. We introduce a Neural Latent-Space Gaussian Process (LGP) architecture for this task and critically evaluate its ability to reconstruct spatial lipid distributions, using held-out mouse brain sections to assess generalization.
We propose a Gaussian Process prior for the distribution of low-dimensional lipid profiles across the mouse brain (with Mat\'ern or a Riemann manifold kernel). This allows us to encode 3D coordinates in a small number of latent 3D volumes that are then mapped to particular lipids using a multilayer perceptron (MLP). We performed an exhaustive analysis over a range of algorithms from simple linear regression, XGBoost, MLP to state of the art models such as Neural Transcriptomic Field, graph-based SPA3D for spatial omics reconstruction and DEEPSPATIAL. We additionally evaluate GPLFR as an alternative Gaussian-process-based approach.  
Across 173 measured lipids, LGP outperforms all evaluated baselines, including recent spatial-omics models and GPLFR, achieving a mean correlation of $0.613 \pm 0.03$ and demonstrating the effectiveness of our approach for sparse 3D lipid interpolation.

\section{Background and problem statement}

Lipids represent most of the brain's dry weight \cite{Bassini2025.10.13.682018} and are critical for synaptic transmission and signal propagation, while non-standard brain lipid composition is associated with common and rare diseases. Bassini et al.~\cite{Bassini2025.10.13.682018} attempt to characterize their spatial organization by building the Mouse Brain Lipid Atlas which represents fine-grained biochemical structures as Lipizones - cell type clusters aligned with the brain's functional anatomy. The lipid profiles are obtained via matrix-assisted laser desorption/ionization mass spectrometry imaging (MALDI-MSI) \cite{norris2013analysis}. The dataset consists of cellular-resolution images across 109 brain sections from 11 mice of 8 weeks of age and maps the distributions of 173 annotated lipids. 

We aim to fully reconstruct the entire 3D lipid distribution for all 173 lipids, beyond the represented brain portion by the available 109 slices. Bassini et al.~\cite{Bassini2025.10.13.682018} used a 3D interpolation algorithm based on radially exponentially-decaying weighted neighbors with the Allen brain regions as anatomical guides which will serve as one of the baseline algorithms that we will compare our results with. We will work on the same dataset as in \cite{Bassini2025.10.13.682018}, which is composed of 2 densely sampled mouse brains of 32 slices each, and 6 sparsely sampled mouse brains (3 male and 3 female) with 6 slices each, that have been mapped to the Allen Mouse Brain reference brain \cite{allen}. We will also use the density reference image and spatial annotations of the Allen Mouse Brain reference in order to reconstruct the lipid profiles at the non-empty voxels.

Each slice of the curated dataset contains between $\sim$31,000 and $\sim$65,000 datapoints \ref{sec:maldi_analysis}, each data point containing the MALDI values of 173 lipids. We will define a number of training data folds where we hold out all the slices of 2 different mice in each fold (with the exception of the densely sampled mice, which are part of all our folds). We validate the quality of the reconstruction by evaluating the mean square error and correlation of our method's output with the measurements of the hold-out sections from the test set.

\begin{figure}
    \centering
    \includegraphics[width=0.75\linewidth]{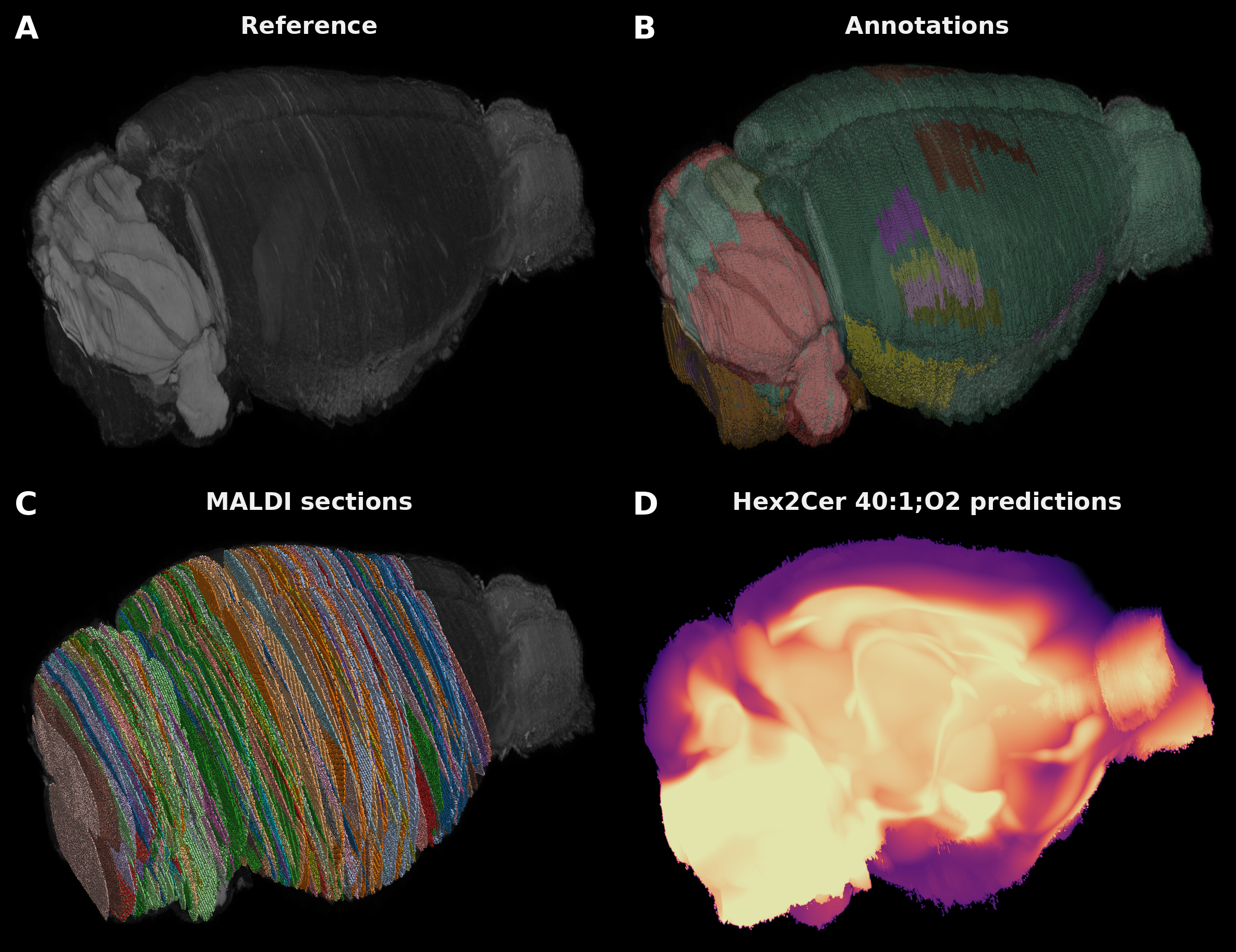}
    \caption{A. The Allen reference brain image. B. The Allen brain mouse annotations. C. The MALDI-MSI image sections. D. Generated Hex2Cer 40;1;O2 3D image }
    \label{fig:overview}
\end{figure}

Specifically, we use a latent-space Stochastic Variational Gaussian Process (SVGP) with the Matérn kernel as the coordinate encoder and a multilayer perceptron (MLP) decoder. The MLP decoder nonlinearly combines the latent Gaussian process (spatial fields) and leverages the correlation structure across lipids. In fact, we analyzed the pairwise correlation between all lipids and found very specific spatial patterns (see \ref{sec:maldi_analysis} details) - $\sim$41\% of the variance can be explained by the first principal component, with a median correlation score $R^2 = 0.4$. Additionally, we attempted to improve the architecture by using an Implicit Manifold Gaussian Process \cite{fichera2023implicit} as a replacement for the Matérn Gaussian Process, with the assumption that the geometry-respecting spectral distance creates a better covariance function that respects the brain's anatomical boundaries. We provide more mathematical background and an illustrative example in the Appendix \ref{sec:math_background}. An Implicit Manifold Gaussian Process works best when the data lies on a lower dimensional manifold that the covariance function implicitly captures via the spectral distance imposed by the k-NN Graph Laplacian, however because the Allen reference brain is very dense in this regard, we have to introduce additional penalties in the kNN cross-region edges that capture the brain's anatomy beyond what the simple geometric k-NN can represent. We annotated each voxel by the top 4 hierarchical annotation levels from Allen \cite{allen} that define 67 anatomical regions. See figure \ref{fig:overview} for an overview of the underlying data, the annotations and an example lipid profile reconstruction.

Additionally, in one particular experiment, we  attempt to replace the MLP with a Gaussian Process (GP) Latent Factor Regressor\cite{stevenson2026gaussian}. We compare the results against a panel of baselines, including a simple mean, a linear model, XGBoost, MLP, and a Graph Neural Network (on the same graph used by the Implicit Manifold Gaussian Process) as well as against a number of recently proposed architectures used for spatial omics reconstruction.

\section{Related work}

Both the spatial omics reconstruction literature and low-data Gaussian process literature concern themselves with the problem of reconstructing dense continuous biological molecular structures from dense and noisy discrete measurements. The lipid reconstruction task is very similar in this regard, as mass spectrometry imaging has many lipid channels registered to the same discrete voxels from which we want to recover underlying patterns to be able to predict the missing lipid values in the whole brain.

Early approaches to 3D spatial omics treated reconstruction as an alignment problem: consecutive 2D sections are registered into a common frame and stacked. PASTE \cite{zeira2022alignment} is the canonical method, offering pairwise alignment and a "center-slice" mode. PASTE works by using Fused Gromov-Wasserstein optimal transport for a slice to slice matching problem. PASTE assumes that inter-slice distances are near-identical, which motivated a range of further research into graph-based methods such as SpaGCN \cite{hu2021spagcn}, GraphST \cite{long2023spatially}, BayesSpace \cite{zhao2021spatial} and a Deep GP \cite{jones2023alignment} for alignment.

More recent research related to 3D lipid reconstruction of biological structures fits a continuous global field as opposed to slice-to-slice interpolation.  SPA3D (\cite{tang20263d}) builds a 3D adjacency graph encoding real $z$ distances, applies an anti-leakage Fourier transform for denoising, and refines a graph convolutional network (GCN) embedding with deep embedded clustering.  Neural Transcriptomic Field (NTF) (\cite{gong2026ultra}) fits an implicit neural field with multi-resolution hash-grid encoding, denoising by construction, and super-resolving from $\sim$10 slices. DEEPSPATIAL (\cite{yang2026reconstructing}) reframes reconstruction as optimal-transport flow matching, with a Transformer parameterizing a probability-flow ordinary differential equation (ODE) sampled at any depth. These are validated on tissues including the whole-mouse-brain atlas (\cite{zhang2023molecularly}).  Our Latent Gaussian Process (either Mat\'ern kernel or implicit manifold kernel (\cite{fichera2023implicit})) can be viewed as a method to produce a continuous smooth field between slices.

Low-data noisy regression has been tackled by various GP-related research (\cite{alvarez2012kernels}) and multi-output is usually done via coregionalization. There are extensions like LV-MOGP (\cite{dai2017efficient}) where output similarity is evaluated via a kernel based on embeddings, and further approximate variations on that (\cite{hensman2013gaussian}, \cite{bruinsma2020scalable}). It is also common to use PCA to reduce the output before individual regressions (\cite{higdon2008computer}). GPLFR (\cite{stevenson2026gaussian}) is more directly aligned with our method: it decodes a prior latent state of GP through a marginalized linear decoder. The linear decoder can be directly compared against our MLP decoder and, in this paper, we will evaluate the additional gain of the decoder on top of an individual per lipid MLP as well as the usage of the graph-Laplacian eigenmodes (prediction-aware, tissue-intrinsic) over a PCA basis, and to the mode-budget/node-density tradeoff. 

The precursor to \cite{fichera2023implicit} is the manifold GP of \cite{calandra2016manifold} — which 
learns a mapping for the input space so that we can apply a stationary kernel. The Implicit Manifold Gaussian Process uses an approximation of the manifold via a kNN graph construction where the distance between nearest neighbors is approximated as the Euclidean, but the kernel itself is based on the eigenfunctions of the Laplace-Beltrami operator.

\section{Methodology}

\begin{figure}[htbp]
    \centering
    \includegraphics[width=1.0\linewidth]{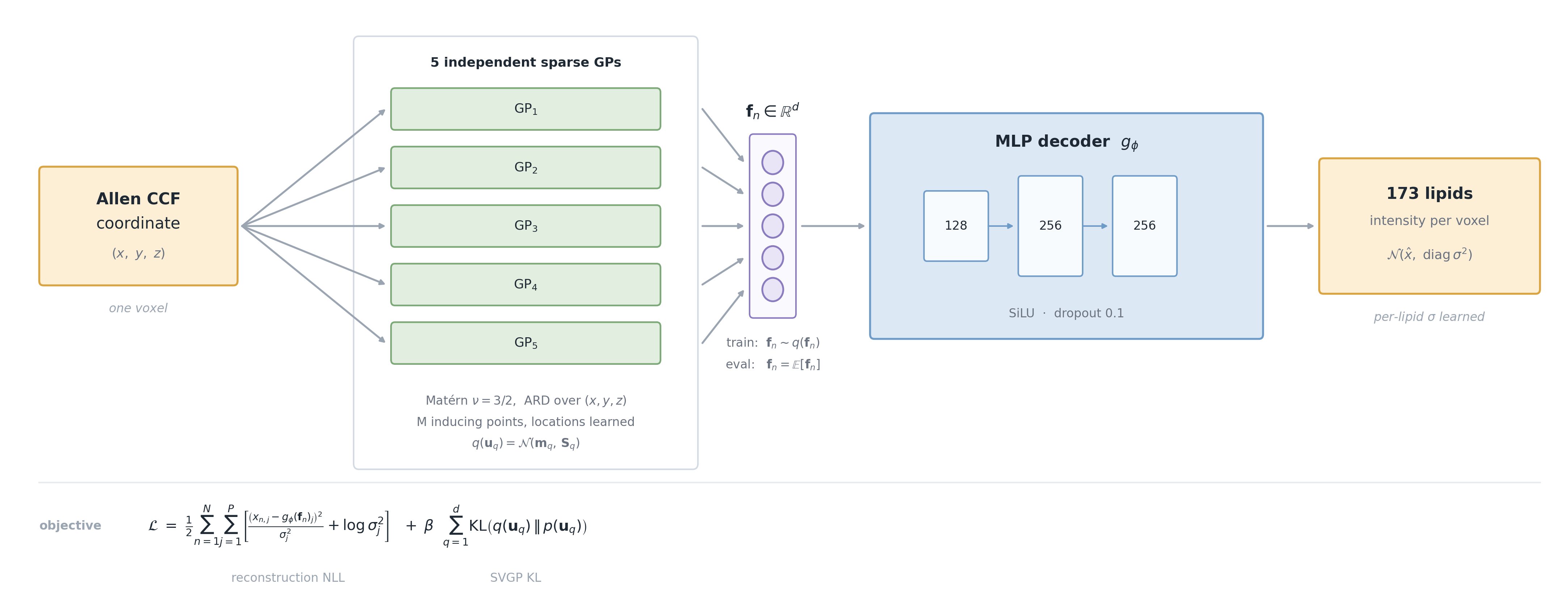}
    \caption{Latent Gaussian Process Model architecture}
    \label{fig:architecture}
\end{figure}

 We propose a Latent Gaussian Process (LGP) architecture for the reconstruction of lipid profiles from spatial Allen Common Coordinate Framework (CCF) coordinates.  Throughout this paper, \textbf{LGP} refers to the proposed Latent Gaussian Process architecture with a Euclidean Matérn kernel, unless specifically clarified otherwise. The overall architecture of the model is represented in Figure \ref{fig:architecture}.

Our primary focus is a comparison of the usage of LGP  against a broad baseline: Linear interpolation, an MLP (including one that uses the eigenvector projection as a feature), a Graph Neural Network (including one trained on the same graph as the Implicit Manifold GP k-NN graph), and against a panel of recent algorithms: SPA3D \cite{tang20263d}, NTF \cite{gong2026ultra}, DEEPSPATIAL \cite{yang2026reconstructing} and GPLFR \cite{stevenson2026gaussian}. Our secondary objective is to evaluate whether an Implicit Manifold GP \cite{fichera2023implicit} can be a suitable replacement for a Matérn GP in our latent decoder.

\subsection{Latent Gaussian Process}
 
Both the Latent Gaussian Process with the Matérn Kernel and the Implicit Manifold Gaussian Process use a shared setup: a configurable number of GPs that map the Allen CCF (x, y, z) coordinates to the latent volume(s) that serve as the input to a simple MLP. The presented results use 5 GPs (i.e. a 5 dimensional latent space representing 5 latent volumes) and a 4 layer MLP (256, 256, 128, 173), SiLU activation with a 0.1 dropout. 

We model the latent coordinate function with an SVGP fully specified by a mean function and a covariance (kernel) function. We used a sparse variational approximation with $M\ll N$ ($M=1000$, $N\approx 77 \text{\,million}$) inducing points $Z$ and a variational posterior $q(\mathbf u)=\mathcal N(\mathbf m,\mathbf S)$ to resolve the intractable exact inference on such a large training set. We trained each model configuration on 8 different folds (each fold using 2 different mice as test data), and, in addition to performing a head to head comparison, also broadly compared these to a panel of baselines and state of the art models. We provide detailed mathematical background on both the Stochastic Variational Gaussian Process and Implicit Manifold Gaussian Process in the Appendix, Section \ref{sec:math_background}. 

In short, predictions are obtained by propagating the variational posterior
$q(\mathbf u_d)$ to a new location $\mathbf c_\ast$ through the GP conditional
$p\big(f_d(\mathbf c_\ast)\mid\mathbf u_d\big)$, marginalizing out the inducing
values. The $D$ latent GPs are
mutually independent, each one having its own posterior
$q(\mathbf u_d)=\mathcal N(\mathbf m_d,\mathbf S_d)$, inducing locations
$Z_d$, mean weights $(\mathbf w_d,b_d)$ and kernel hyperparameters
$(\sigma_{f,d},\ell_d)$. For each
latent dimension $d=1,\dots,D$ this yields a Gaussian predictive
$q\big(f_d(\mathbf c_\ast)\big)=\mathcal N\!\big(\mu_{\ast,d},\,\sigma_{\ast,d}^2\big)$
with
$$\mu_{\ast,d}=\underbrace{\mu_d(\mathbf c_\ast)}_{\text{LinearMean}}+\mathbf k_{\ast Z_d}^{(d)}\,\mathbf K_{Z_dZ_d}^{(d)\,-1}\big(\mathbf m_d-\boldsymbol\mu_d(Z_d)\big)$$
$$\sigma_{\ast,d}^2=\underbrace{k_{\ast\ast}^{(d)}}_{\text{prior}}-\underbrace{\mathbf k_{\ast Z_d}^{(d)}\mathbf K_{Z_dZ_d}^{(d)\,-1}\mathbf k_{Z_d\ast}^{(d)}}_{\text{conditioning on }Z_d}+\underbrace{\mathbf k_{\ast Z_d}^{(d)}\mathbf K_{Z_dZ_d}^{(d)\,-1}\mathbf S_d\,\mathbf K_{Z_dZ_d}^{(d)\,-1}\mathbf k_{Z_d\ast}^{(d)}}_{\text{variational uncertainty}}$$

We used a linear mean function $\mu_d(\mathbf x)=\mathbf w_d^\top \mathbf x + b_d$, and the Matérn kernel described by the covariance function
$$k^{(d)}(\mathbf x,\mathbf x')=\sigma_{f,d}^{2}\,\frac{2^{1-\nu}}{\Gamma(\nu)}\!\left(\frac{\sqrt{2\nu}\,r}{\ell_d}\right)^{\!\nu} \mathcal K_{\nu}\!\left(\frac{\sqrt{2\nu}\,r}{\ell_d}\right),\qquad r=\lVert \mathbf x-\mathbf x'\rVert,$$

where $\nu$ is the (fixed) smoothness and $K_\nu$ the modified Bessel function; with automatic relevance determination (ARD) the single $\ell$ is replaced by per-axis lengthscales $(\ell_1,\ell_2,\ell_3)$. All kernel matrices are Gram matrices of this covariance function evaluated at the relevant point sets: $[\mathbf K_{ZZ}]_{ij}=k(\mathbf z_i,\mathbf z_j)$ over the inducing locations, $[\mathbf k_{\ast Z}]_i=k(\mathbf c_\ast,\mathbf z_i)$ between a test point and the inducing set, and $k_{\ast\ast}=k(\mathbf c_\ast,\mathbf c_\ast)$;  therefore, they inherit the kernel hyperparameters $\sigma_f^2$ and $\ell$, which is how those hyperparameters enter both the predictive equations and the Kullback--Leibler (KL) term. We provide a short description of the Riemann manifold kernel of the Implicit Manifold GP in the next subsections \ref{sec:manifold_gp_explanation} and \ref{sec:spectral_gp_explanation}.

All parameters are learned jointly by minimizing the loss below via the Adam optimizer: the variational parameters $\{\mathbf m_d,\mathbf S_d\}_{d=1}^{D}$ (with $\mathbf S_d$ parametrized by its Cholesky factor $\mathbf S_d = \mathbf L_d \mathbf L_d ^{\top}$ to remain positive-definite) and inducing locations $\{Z_d\}_{d=1}^{D}$; the kernel hyperparameters $\sigma_{f,d}^2$ and $\ell_d$ (or the ARD lengthscales $\ell_{d,1:3}$); the mean-function weights $\mathbf w_d,b_d$; the per-lipid observation noises $\tau_j^2$; and the decoder weights $\phi$. The smoothness $\nu$ is fixed. The training objective is the negative evidence lower bound (ELBO), a reconstruction term plus a KL regularizer (up to a constant):
$$\mathcal{L}
=\underbrace{\frac12\sum_{n=1}^{N}\sum_{j=1}^{P}\!\left[\frac{\big(x_{n,j}-g_\phi(\mathbf f_n)_j\big)^2}{\tau_j^2}+\log\tau_j^2\right]}_{\text{Monte Carlo estimate of NLL} \;=\; -\,\mathbb{E}_{q(\mathbf f)}[\log p(\mathbf x\mid \mathbf f)]}
\;+\;\underbrace{\sum_{d=1}^{D}\mathrm{KL}\!\big(q(\mathbf u_d)\,\|\,p(\mathbf u_d)\big)}_{\text{SVGP KL}}$$

The reconstruction term is the Gaussian negative log-likelihood of the observed lipids $x_{n,j}$ under the decoder output $g_\phi(\mathbf f_n)_j$, where the latent $\mathbf f_n$ is drawn from the variational predictive $q(\mathbf f)$ and decoded by the MLP $g_\phi$. Because $g_\phi$ is nonlinear, the expectation $\mathbb{E}_{q(\mathbf f)}[\log p(\mathbf x\mid\mathbf f)]$ has no closed form and is estimated by Monte Carlo with a single reparameterized sample $\mathbf f=\boldsymbol\mu_\ast+\boldsymbol\Sigma_\ast^{1/2}\boldsymbol\epsilon$, $\boldsymbol\epsilon\sim\mathcal N(\mathbf 0,\mathbf I)$, per step; at evaluation the posterior mean $\mathbf f_n=\boldsymbol\mu_\ast(\mathbf c_n)$ is used instead. The KL term, summed over the $D$ independent latent GPs, keeps each variational posterior close to its prior, and is available in closed form.

As an implementation detail, rather than parametrizing $\mathbf u_d$ directly, we write $\mathbf u_d=\boldsymbol\mu_d(Z_d)+\mathbf L_d\mathbf v_d$, where $\mathbf L_d$ is a Cholesky factor of $\mathbf K^{(d)}_{Z_dZ_d}$, and learn the variational distribution of $\mathbf v_d$. The KL then reduces to
$\mathrm{KL}\big(q(\mathbf v_d)\,\|\,p(\mathbf v_d)\big)=\tfrac12\!\left[\operatorname{tr}(\mathbf S_d)+\mathbf m_d^\top\mathbf m_d-M-\log|\mathbf S_d|\right],$
where $\mathbf m_d,\mathbf S_d$ are the variational mean and covariance of $\mathbf v_d$. Equivalently, in unwhitened coordinates the same quantity is $\tfrac12\big[\operatorname{tr}(\mathbf K^{-1}\mathbf S)+(\mathbf m-\boldsymbol\mu_d(Z_d))^\top\mathbf K^{-1}(\mathbf m-\boldsymbol\mu_d(Z_d))-M+\log(|\mathbf K|/|\mathbf S|)\big)$ with $\mathbf K=\mathbf K^{(d)}_{Z_dZ_d}$; the two parametrizations give the same ELBO.

\begin{figure}[htbp]
    \centering
    \includegraphics[width=0.6\linewidth]{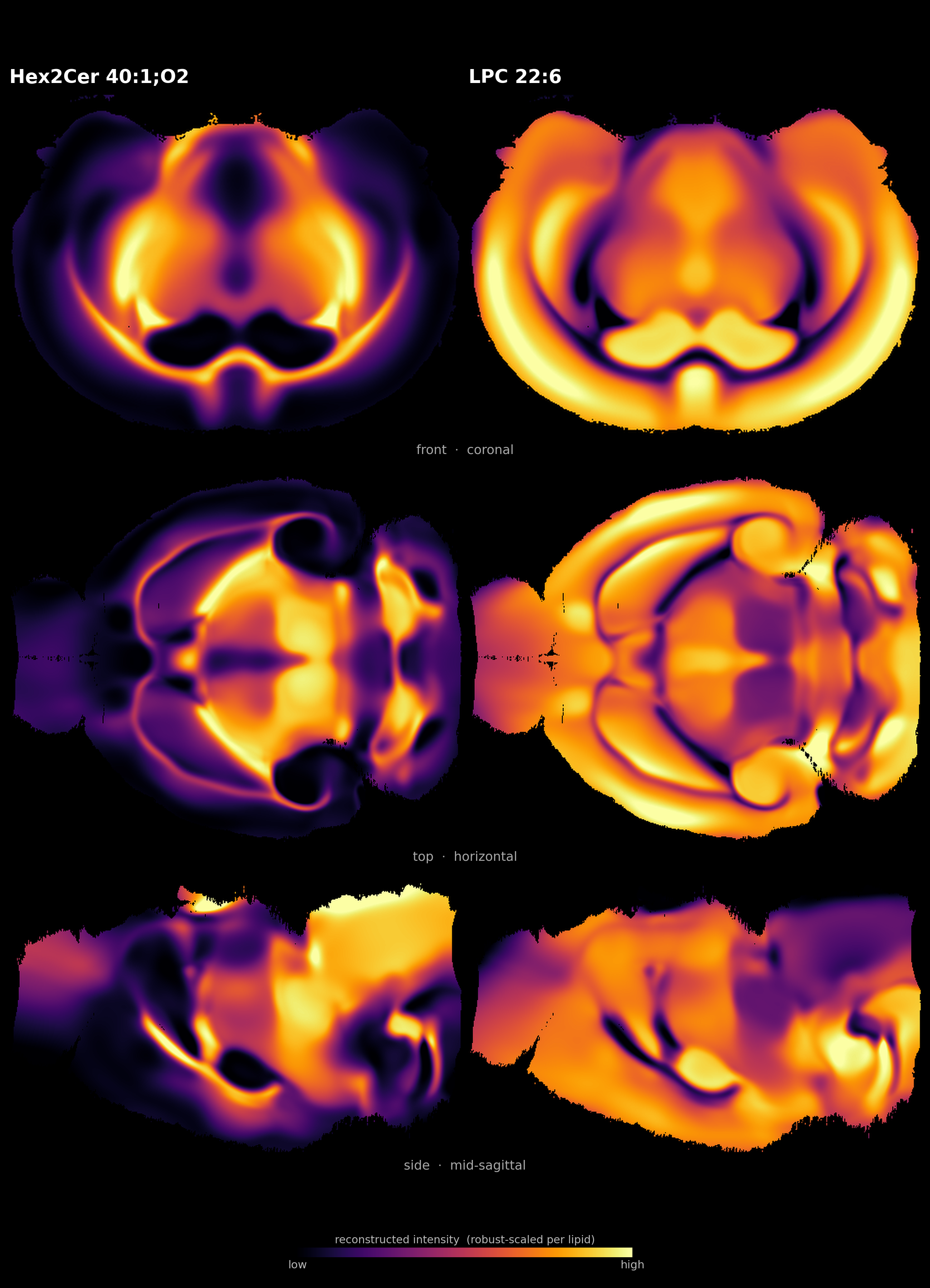}
    \caption{A full scale render using LGP with the Mat\'ern kernel}
    \label{fig:full_scale_render}
\end{figure}

 For a few example 3D reconstruction results see Figure \ref{fig:full_scale_render}.

\FloatBarrier

\subsection{Manifold Gaussian Process}
\label{sec:manifold_gp_explanation}

In the Implicit Manifold Gaussian Process method \cite{fichera2023implicit}, we represent the manifold via a k-NN graph over the point cloud of the data spatial coordinates and assign $$w_{ij} = \exp\!\Big(-\tfrac{\lVert x_i - x_j\rVert^2}{4\varepsilon^2}\Big), \qquad d_i=\sum_j w_{ij}.$$ This is subject to further normalization in order to correct the potential non-uniform graph node sampling:
$$\widetilde w_{ij}=\frac{w_{ij}}{d_i\,d_j},\qquad \widetilde d_i=\sum_j \widetilde w_{ij}.$$ In the original paper, the graph bandwidth $\epsilon$ is a learnable parameter which is computationally intractable in our setup as we would have to recompute the eigenvectors at each backward pass. We decided on a hard-coded constant $\epsilon=0.1$, which was determined via cross-validation, and is able to assign a meaningful weight as to differentiate the nearest neighbors from anatomically different structures. From the normalized affinities $\widetilde{\mathbf W}$ and degrees $\widetilde{\mathbf D}=\mathrm{diag}(\widetilde d_i)$ we form a graph Laplacian $\boldsymbol\Delta$, the discrete approximation of the Laplace–Beltrami operator $\Delta_{\mathcal M}$ of the underlying manifold.

The Riemann Kernel itself is:
$$k(i,j)=\sum_{k} s(\lambda_k)\,\phi_k(i)\,\phi_k(j),\qquad s(\lambda)=\Big(\tfrac{2\nu}{\ell^2}+\lambda\Big)^{-\nu}$$

where $\{\lambda_k,\phi_k\}$ are the eigenvalues and eigenfunctions of the graph Laplacian $\boldsymbol\Delta$, i.e. $\boldsymbol\Delta\phi_k=\lambda_k\phi_k$, and $s(\lambda)$ is a low-pass spectral filter parameterized by the smoothness $\nu$ and lengthscale $\ell$. The eigenfunctions $\phi_k$ can be seen as harmonics of the discrete manifold — smooth global modes for small $\lambda_k$, increasingly oscillatory ones for large $\lambda_k$ — so the kernel is a smoothness prior that damps high-frequency modes on the tissue. Under the density ($\alpha=1$) normalization of the affinity graph, the graph Laplacian converges, as $N\to\infty$ and the bandwidth $\varepsilon\to0$, to the Laplace–Beltrami operator $\Delta_{\mathcal M}$ of the underlying manifold, so that $k$ converges to the exact Riemannian Matérn kernel; crucially, this pointwise convergence holds only for random-walk normalization $\boldsymbol\Delta_{\text{rw}}=\widetilde{\mathbf D}^{-1}\widetilde{\mathbf L}$. We therefore use $\boldsymbol\Delta_{\text{rw}}$ for the covariance.

For a test point $\mathbf x_\ast$ not in the training graph, each eigenfunction is extended by Nyström interpolation from its $k$ nearest graph nodes $\mathcal N(\mathbf x_\ast)$:
$$\phi_k(\mathbf x_\ast)=\rho\big(d_{\text{nn}}\big)\sum_{j\in\mathcal N(\mathbf x_\ast)}\widetilde w_{\ast j}\,\phi_k(\mathbf c_j),\qquad
\widetilde w_{\ast j}=\frac{w_{\ast j}}{\textstyle\sum_{i}w_{\ast i}},\quad
w_{\ast j}=\exp\!\Big(-\tfrac{\lVert\mathbf x_\ast-\mathbf c_j\rVert^2}{4\varepsilon^2}\Big),$$
Nyström interpolation is a normalized affinity-weighted average of neighbor eigenvectors, gated by a smooth bump $\rho(d)=\texttt{bump}(d;\,\alpha,\beta)$ of the distance $d_{\text{nn}}$ to the nearest node, with support radius $\alpha=\text{bump\_scale}\cdot\varepsilon$ and decay $\beta=\text{bump\_decay}$:
$$\rho(d)=
\begin{cases}
\displaystyle\exp\!\left(\dfrac{\beta}{\alpha^{2}}-\dfrac{\beta}{\alpha^{2}-d^{2}}\right) & d<\alpha,\\[3mm]
0 & d\ge\alpha.
\end{cases}$$
 Points outside the manifold beyond $\text{bump\_scale}\cdot\varepsilon$ do not contribute covariance.

In \cite{fichera2023implicit}, the graph is constructed from the closest $k$-NN neighbors, under the fundamental assumption that, locally, the geodesic distance between two connected voxels is well approximated by their Euclidean distance. However, as shown in the figure below, this construction yields a very dense graph: because $k$-NN connects spatially adjacent voxels regardless of anatomy, edges short-circuit across region boundaries, and the graph geodesic between any two voxels collapses toward their ambient Euclidean distance — defeating the purpose of the Implicit Manifold GP, which is to respect anatomical boundaries and surfaces. We therefore alter the graph using the Allen atlas labels $\ell(\cdot)$ of the voxels. First, we introduce a soft penalty on any edge whose endpoints lie in different regions, inflating its squared distance by a factor $\gamma>1$,

$$ \tilde d_{ij}^{\,2}=\begin{cases}\gamma\,\lVert\mathbf c_i-\mathbf c_j\rVert^2 & \ell(i)\neq\ell(j)\\[2pt]\lVert\mathbf c_i-\mathbf c_j\rVert^2 & \ell(i)=\ell(j)\end{cases}
\qquad
 w_{ij}=\exp\!\Big(-\tfrac{\tilde d_{ij}^{\,2}}{4\varepsilon^2}\Big),
$$

so that a cross-region affinity is inflated super-exponentially, $w_{ij}\mapsto w_{ij}^{\gamma}$. Because we observed various cross-region noisy voxels, we first denoise the labels by iterated majority vote — each voxel adopts the most common label among its graph neighbors — so that isolated mislabeled voxels do not create spurious boundaries. Finally, we optionally apply a hard prune, removing a fraction of the remaining cross-region edges outright (in a connectivity-preserving manner, retaining each node's strongest incident edge as a bridge so the graph does not fragment), which severs the residual short-circuits entirely.

\subsection{Spectral Manifold Gaussian Process}
\label{sec:spectral_gp_explanation}
We ran a separate experiment of assigning individual eigenfunctions their own independent weights. We represent the function as a biased weighted sum of eigenbases,
$$f(\mathbf x)=\sum_{k=1}^{K} w_k\,\phi_k(\mathbf x)\;+\;\mathbf w_{\text{lin}}^\top\mathbf x+b,$$
the manifold harmonics $\phi_k$ (extended off-graph by the Nyström map of the previous subsection) serve as basis functions, and the only latent variables are the mode weights $\mathbf w=(w_1,\dots,w_K)$. The Matérn prior enters through the spectral density: each weight is given an independent Gaussian prior $p(\mathbf w)=\mathcal N\!\big(\mathbf 0,\,\sigma^2\boldsymbol\Phi\big)$ with $\Phi_k=s(\lambda_k)=(2\nu/\ell^2+\lambda_k)^{-\nu}$, so that smooth low-frequency modes are favored and oscillatory ones suppressed. We approximate the posterior over the weights with a diagonal Gaussian $q(\mathbf w)=\mathcal N(\boldsymbol\mu_w,\operatorname{diag}\mathbf s_w)$, initialized at the prior so the KL begins near zero. Predictions require no kernel matrix or inverse: interpolating the eigenfunctions to a query point ($\mathbf U_\ast$) gives the closed-form predictive mean and variance directly as $\mu(\mathbf x_\ast)=\mathbf U_\ast\boldsymbol\mu_w+\mathbf w_{\text{lin}}^\top\mathbf x_\ast+b$ and $\operatorname{var}(\mathbf x_\ast)=\mathbf U_\ast^{\odot2}\mathbf s_w$. Training minimizes the same negative ELBO as the inducing-point model — a decoder reconstruction term plus $\beta$ times the KL — but here the KL is a closed-form diagonal Gaussian divergence in spectral space.

\subsection{Baseline models}

We implemented a broad set of baselines to compare the Latent Gaussian Process architecture against, with the best model configuration of each category chosen via cross-validation across the list of hyperparameters described in \ref{sec:hyperparameters}, and the winning configuration in Table \ref{tab:winning-configs}:

\paragraph{Mean} A simple mean that defines the null floor.

\paragraph{Linear} - A linear ridge regression on the coordinates $\hat y_j(\mathbf c)=\mathbf w_j^\top \mathbf c + b_j$, $\mathbf c=(x,y,z)$, with one independent regressor per lipid.

\paragraph{XGBoost} -  One independent gradient-boosted regressor per lipid (
$400$ trees, max depth $6$, early stopping). A strong non-spatial tabular
baseline that captures nonlinear coordinate--lipid relationships without any
notion of manifold geometry.

\paragraph{MLP} -  A coordinate$\rightarrow$lipid multilayer perceptron
($256\text{--}256\text{--}128$, SiLU, dropout $0.1$). Its architecture is
identical to the GP decoder, so the ``MLP vs GP'' gap isolates the
contribution of the latent Gaussian-process prior over and above the same
network.

\paragraph{MLP eigen} - The MLP applied to the concatenation of the original coordinates and the Laplacian eigenmap features. It tests whether the manifold spectral basis alone is useful,
decoupled from the GP.

\paragraph{GCN} - A graph convolutional network ($512\text{--}512\text{--}256$,
dropout $0.1$) over a $k$-nearest-neighbor graph of the coordinates ($k=15$,
inverse-distance read-out with $8$ neighbors). It is the deep geometric
counterpart to the manifold GP, operating on the same neighborhood graph
($k=15$) but with a learned message-passing prior instead of a spectral one.

\paragraph{EUCLID} The original implementation.
EUCLID~\cite{Bassini2025.10.13.682018} is the analysis pipeline released with the
Lipid Brain Atlas, covering normalization, Lipizone clustering and
differential analysis. The anatomical interpolation operates per lipid on the Allen CCF grid: measured pixels are log-transformed and averaged within each voxel, giving $s_v$ defined on the set $\mathcal D$ of voxels containing at least one measurement.
Every voxel then receives an inverse-distance weighted mean of the measured
voxels that share its Allen label,
\begin{equation}
\hat y_v=\frac{\sum_{u\in\mathcal N(v)}e^{-\lVert u-v\rVert}\,s_u}
              {\sum_{u\in\mathcal N(v)}e^{-\lVert u-v\rVert}},
\qquad
\mathcal N(v)=\big\{\,u\in\mathcal D:\ \lVert u-v\rVert\le R,\ \ \ell_u=\ell_v\,\big\},
\label{eq:euclid}
\end{equation}
with $\ell_v$ the Allen label at $v$, $\lVert\cdot\rVert$ the Euclidean distance
in voxel units and $R=26$ voxels ($2.6$\,mm). Voxels with no admissible donor
fall back to an unweighted $10^3$ box mean. We used both the original $100\,\mu$m Allen CCF grid and provided Allen annotations as well as the $25\,\mu$m reference image, and used the best cross-validation result as the baseline.

\subsection{State of the art models}
We further compare against four recent methods for reconstructing continuous
molecular fields from sparse tissue sections, adapted to the MALDI lipid task
($\texttt{predict(coords)}\rightarrow\texttt{lipids}$ over the dense voxel grid,
with ``section'' defined by binning the coronal $x_{\text{ccf}}$ axis). We evaluated these models using the configurations defined in \ref{sec:hyperparameters}, and the winning configuration in Table \ref{tab:winning-configs} chosen via cross-validation.

\paragraph{NTF \cite{gong2026ultra}.} An implicit neural field with a
multi-resolution hash-grid encoding ($12$ levels, $8$ features/level, hashmap
$2^{19}$, resolutions $16\rightarrow1024$) decoded by an MLP
($256\text{--}256\text{--}128$, matched to ours). It adds a total-variation
smoothness prior ($0.05$), a Monte-Carlo point-spread-function average
($4$ samples, $\sigma=0.01$) that reconstructs the deblurred field, and a
heteroscedastic per-slice bias network. Included as the leading implicit-neural
reconstruction method.

\paragraph{SPA3D \cite{tang20263d}.} A spatial-pattern-enhanced, $z$-aware
graph convolutional network. Each section is denoised by an anti-leakage Fourier
transform (keeping $50\%$ of coefficients), and a GCN
($512\text{--}512\text{--}256$) operates on a global Gaussian graph ($k=15$,
$8\times10^4$ nodes) whose metric explicitly scales the true inter-slice distance
along $x_{\text{ccf}}$ ($z$-weight $0.3$). Included as the graph-based method
that, unlike PASTE, uses real $z$-distances.

\paragraph{DEEPSPATIAL \cite{yang2026reconstructing}.} The official flow-matching
reconstruction: unbalanced optimal-transport correspondences (entropic
regularization $0.3$, relaxation $\tau=0.05$) between real coronal sections, a
transformer velocity field (hidden $256$, depth $6$, $8$ heads) trained by flow
matching, and probability-flow ODE synthesis ($50$ steps) of the tissue between
measured sections (cross-mouse pairing). Run faithfully via the official package
rather than shoehorned into the regression harness, as the paper's
cell-resolution, generative formulation intends.

\paragraph{GPLFR \cite{stevenson2026gaussian}.} Gaussian-process latent factor
regression: a low-dimensional latent GP ($d=8$, $256$ inducing points, Mat\'ern
$\nu=5/2$) with an analytically marginalized linear decoder. We evaluate it on
our manifold (Riemann) basis, so it also serves as a test of the marginalized
linear decoder against our nonlinear one. Included as the most closely related
GP-based method.

\FloatBarrier

\section{Results}

\begin{figure}[ht]
    \centering
    \includegraphics[width=1.0\linewidth]{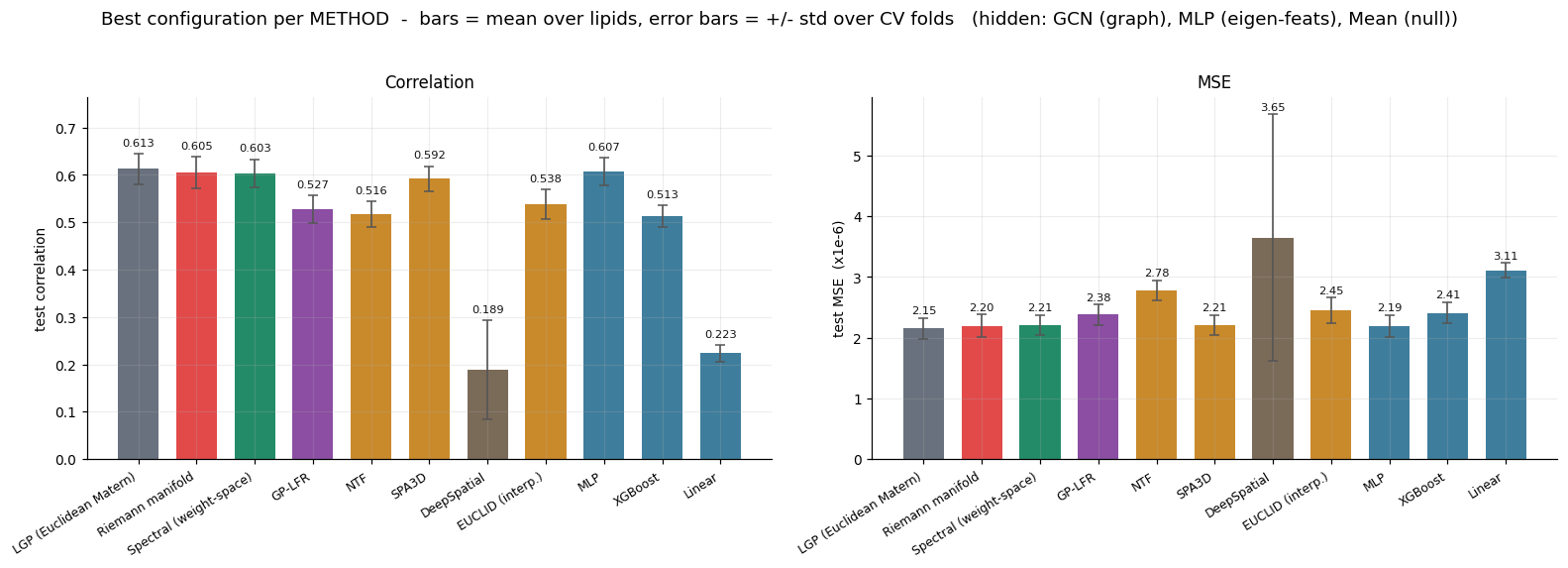}
    \caption{Ranked model correlation summary}
    \label{fig:model_corr_stacked}
\end{figure}

\begin{figure}[ht]
    \centering
    \includegraphics[width=1.0\linewidth]{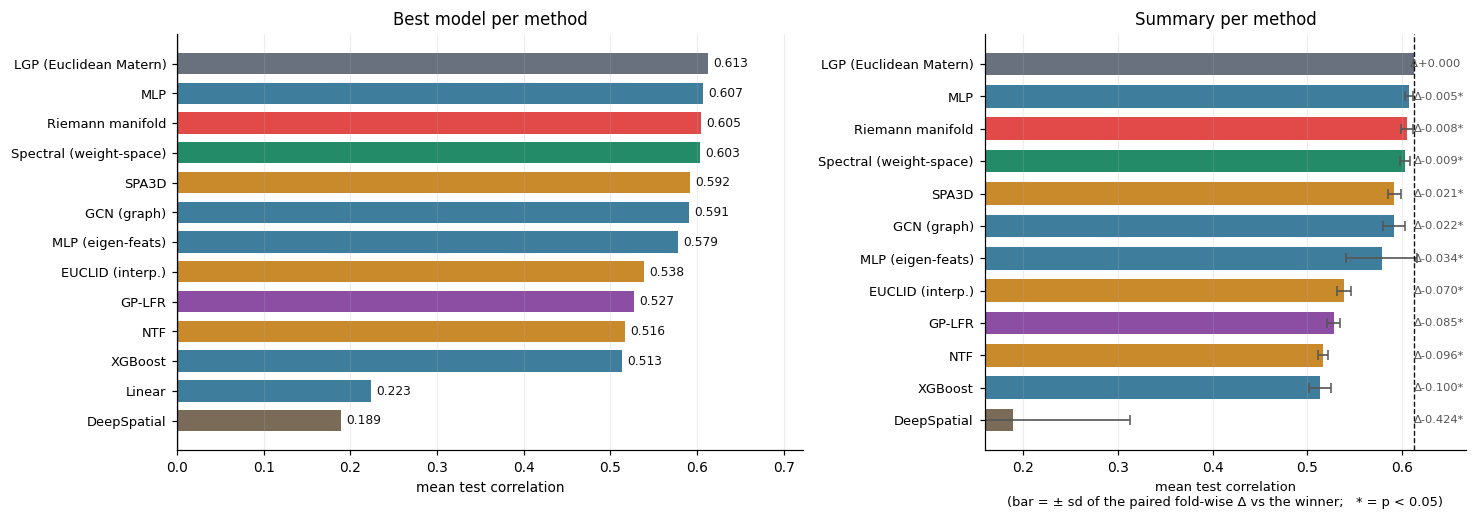}
    \caption{Model MSE and correlation comparison}
    \label{fig:model_corr_comparison}
\end{figure}

We ran all our experiments across 8 folds with the exhaustive list of attempted hyperparameters presented in the Appendix Section \ref{sec:hyperparameters} (Tables \ref{tab:hyperparams} and \ref{tab:models}). The winning configuration \ref{tab:winning-configs} is the Latent Gaussian Process using the Matérn kernel with $\nu=1.5$ and the inducing points being data-derived, achieving a mean correlation of 0.613 $\pm0.03$ (Figures \ref{fig:model_corr_stacked} and \ref{fig:model_corr_comparison}). Table \ref{tab:winning-configs} presents the ranked detailed results as well as the parameters of the best model, with table \ref{tab:config-legend} offering a legend for interpreting the configuration parameters.

The resulting reconstructed images \ref{fig:combined_gps} are smooth with the visual artifacts reflecting the underlying geometrical structure that the GP assumes. Neither the Riemann Kernel nor the Spectral Kernel helped the LGP results - the underlying manifold is very dense and the existing annotations are not providing a benefit exploitable by the Riemann Manifold, as seen below in \ref{sec:matern_vs_manifold}.

It's remarkable that a simple regularized Multi-Layer Perceptron offers a very strong baseline with a  correlation of 0.607 $\pm0.029$, with many state of the art models unable to surpass it. Compared to LGP \ref{fig:combined_gps}, we can observe that the MLP shows a number of anomalous non-smooth artifacts \ref{fig:combind_baselines}  (e.g. in the hypothalamus of the Hex2Cer 40;1;O2 lipid). It is also remarkable that the addition of the eigenvector projection, labeled "MLP-eigen", did not improve the MLP results, further indicating that the k-NN graph and its Laplacian do not contribute meaningfully to the reconstruction capability of the lipid profiles.

\FloatBarrier

The EUCLID 3D interpolation \cite{Bassini2025.10.13.682018} achieved a correlation score of 0.538\ $\pm$0.031, surpassing XGBoost, NTF \cite{gong2026ultra} and GPLFR \cite{stevenson2026gaussian} but presented a significant amount of non-smooth visual artifacts \ref{fig:combind_baselines}.

Among the state of the art methods, SPA3D \cite{tang20263d} offers the best correlation, on par with the baseline GCN, resulting in a correlation score of 0.592 $\pm0.027$ and 0.591 $\pm0.044$ respectively. A GCN reuses the same k-NN graph as the Implicit Manifold GP. A visual inspection of either of these models in \ref{fig:combind_sota} and \ref{fig:combind_baselines} shows a strong presence of visual non-smooth artifacts showing that the GCN graph neighbors don't correlate as strongly with particular lipid values as much as the closest neighbors in the Euclidean space - something that neither of these models was fully equipped to exploit.

NTF \cite{gong2026ultra} multi-resolution hash-grid encoding buys it nothing on the MALDI dataset: Its correlation is on par with XGBoost (0.516 vs 0.513), but it retains only 63\% of the R² implied by that correlation — against 93–96\% for every other method — indicating a per-section level/scale error rather than a shortfall in its representation capacity. Sweeping the top grid resolution over 128–512 and the total-variation weight over 0–2.0 moves test correlation by less than 0.001, so the fine levels are effectively unconstrained by the sparse coronal sampling. 

DEEPSPATIAL \cite{yang2026reconstructing} establishes cross-section voxel correspondences by unbalanced optimal transport, learns a flow-matching velocity field over position and lipid intensity, and synthesizes the tissue between measured sections by integrating a probability-flow ODE. It attains the lowest mean correlation of any method (0.189), though its fold-to-fold variance is large enough that it is not reliably separated from the linear baseline (paired $\Delta = -0.034 \pm 0.100$, $t = -0.98$; it exceeds the linear baseline in 3 of 8 folds). Its strongly negative $R^2$ reflects reconstructions that are badly mis-scaled rather than merely uninformative. The cause is the sectioning density. DEEPSPATIAL's UOT step assumes that adjacent sections sample near-identical cell populations, as holds for the micron-scale spacing of the serial-section transcriptomic data it was developed on. Our dataset contains only 4--6 coronal sections per mouse brain each (excluding the two densely sectioned atlas brains), spaced 0.15--2\,mm apart, so consecutive sections are anatomically distinct and the transport coupling has no reliable correspondence to recover. We expect DEEPSPATIAL to be competitive on serially sectioned tissue where the inter-section gap is comparable to the in-plane resolution. Figure \ref{fig:combind_sota} illustrates this.

GPLFR \cite{stevenson2026gaussian} shares our latent GP backbone but replaces the nonlinear decoder with an analytically marginalized linear-Gaussian map, and is run through the identical data pipeline, CV splits and reconstruction path. It reaches a correlation of 0.527  $\pm 0.029$, and its predictions are as well calibrated as the leading methods (attaining 95\% of the $R^2$ implied by its correlation), so the deficit reflects fit quality rather than any systematic bias. We do note that the GPLFR configuration inherits the reference implementation's defaults (256 fixed inducing points, isotropic lengthscales, a single homoscedastic noise scalar) against our tuned configuration (2000 learned inducing locations, per-axis ARD, per-lipid noise), so the gap bounds rather than isolates the cost of a linear decoder. 

\begin{table}[htbp]
  \centering
  \small
  \makebox[\textwidth][c]{\begin{tabular}{llrrr}
        \toprule
        Method & Winning configuration & Corr & $R^2$ & RMSE ($\times10^{-3}$) \\
        \midrule
        \textbf{LGP} & \texttt{\scriptsize MEAN, data, M2000, learnind, ard} & \textbf{0.613\,$\pm$\,0.032} & 0.375\,$\pm$\,0.045 & 0.598\,$\pm$\,0.036 \\
        MLP & \texttt{\scriptsize bs256} & 0.607\,$\pm$\,0.029 & 0.370\,$\pm$\,0.041 & 0.602\,$\pm$\,0.036 \\
        Manifold GP & \texttt{\scriptsize K100, faiss\_atlas\_weighted, x50, randomwalk, learndiff1.0, blend0.8} & 0.605\,$\pm$\,0.033 & 0.365\,$\pm$\,0.046 & 0.603\,$\pm$\,0.038 \\
        Spectral GP & \texttt{\scriptsize K300, faiss\_cluster\_weighted, x50, randomwalk, clk64, sw1.0, cs0} & 0.603\,$\pm$\,0.030 & 0.361\,$\pm$\,0.043 & 0.605\,$\pm$\,0.035 \\
        SPA3D & \texttt{\scriptsize ZW0.5, SPE=none} & 0.592\,$\pm$\,0.027 & 0.351\,$\pm$\,0.035 & 0.609\,$\pm$\,0.032 \\
        GCN & \texttt{\scriptsize atlas\,$\times$50, rootdissolve, dn3, prune0.97} & 0.591\,$\pm$\,0.044 & 0.341\,$\pm$\,0.064 & 0.615\,$\pm$\,0.046 \\
        MLP-eigen & \texttt{\scriptsize NM300, rootdissolve, dn3, prune0.97} & 0.579\,$\pm$\,0.067 & 0.297\,$\pm$\,0.146 & 0.633\,$\pm$\,0.078 \\
        EUCLID & \texttt{\scriptsize $w{=}0$, level\_15annot gate} & 0.538\,$\pm$\,0.031 & 0.290\,$\pm$\,0.039 & 0.643\,$\pm$\,0.035 \\
        GPLFR & \texttt{\scriptsize euclidean, d8} & 0.527\,$\pm$\,0.029 & 0.280\,$\pm$\,0.036 & 0.640\,$\pm$\,0.031 \\
        NTF & \texttt{\scriptsize RES128, TV0.5} & 0.516\,$\pm$\,0.028 & 0.182\,$\pm$\,0.032 & 0.684\,$\pm$\,0.029 \\
        XGBoost & \texttt{\scriptsize bs256} & 0.513\,$\pm$\,0.023 & 0.264\,$\pm$\,0.026 & 0.646\,$\pm$\,0.029 \\
        Linear & \texttt{\scriptsize ---} & 0.223\,$\pm$\,0.018 & 0.047\,$\pm$\,0.008 & 0.739\,$\pm$\,0.019 \\
        DEEPSPATIAL & \texttt{\scriptsize within-mouse, REG0.3} & 0.189\,$\pm$\,0.104 & $-15.3$\,$\pm$\,29.8 & 0.916\,$\pm$\,0.355 \\
        \midrule
        MEAN (null) & \texttt{\scriptsize ---} & --- & $-0.010$\,$\pm$\,0.005 & 0.755\,$\pm$\,0.020 \\
        \bottomrule
      \end{tabular}
  }
  \caption{Best configuration per method, ranked by mean test correlation over
    173 lipids $\times$ 8 CV folds. Uncertainties are standard deviations across
    the 8 folds. The constant \textsc{mean} predictor has undefined correlation.}
  \label{tab:winning-configs}
\end{table}

\FloatBarrier
\subsection{Matérn vs manifold kernel}
\label{sec:matern_vs_manifold}

The Allen reference mouse brain model is very dense \ref{fig:overview} - within this structure the distance between most pairs of voxels can be well approximated by the Euclidean distance, coupled with the eigenvector mode truncation is not expected to provide a meaningful improvement.

\begin{figure}[htbp]
    \centering
    \includegraphics[width=0.8\linewidth]{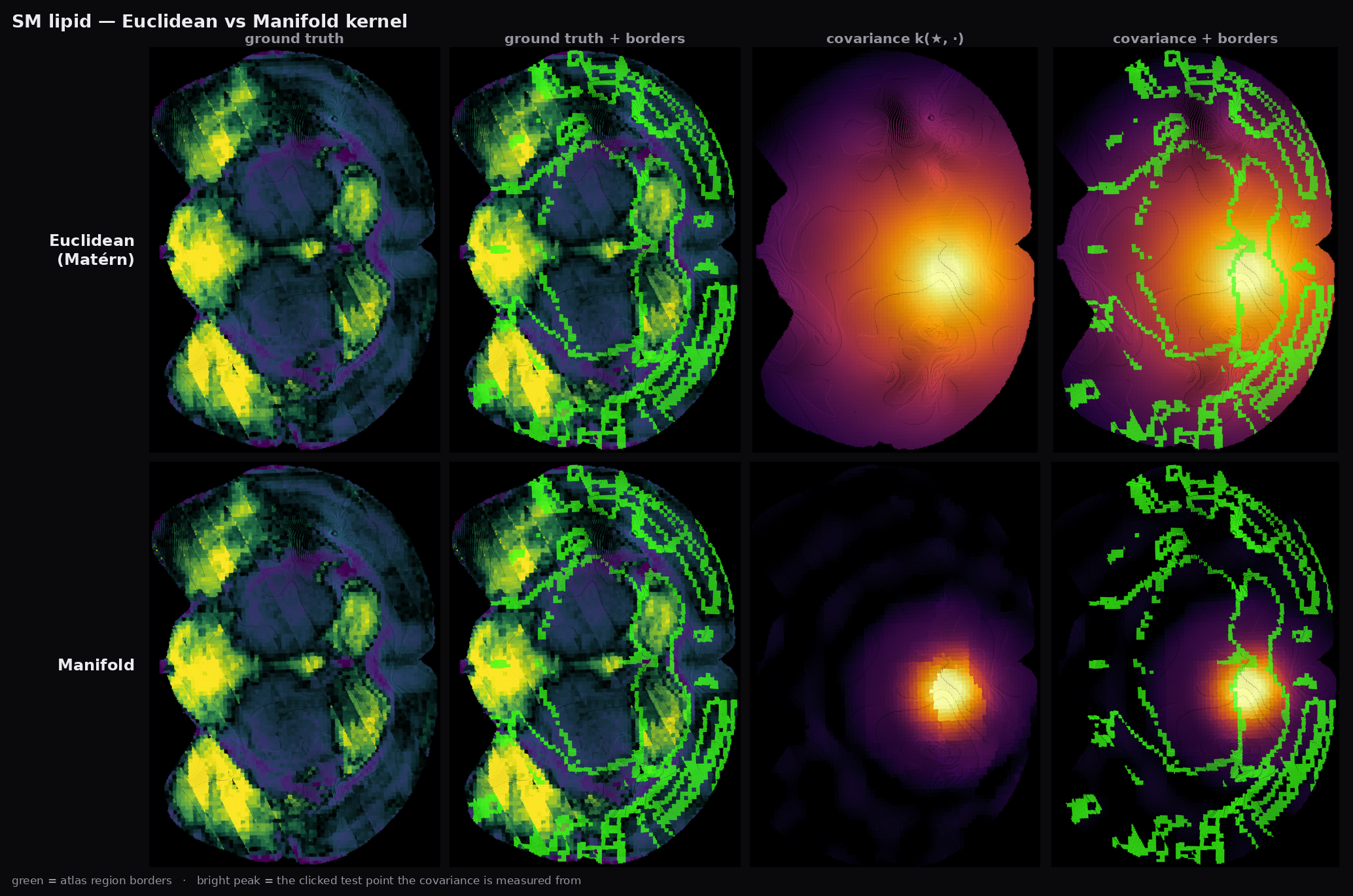}
    \caption{Manifold kernel vs Euclidean Matérn kernel}
    \label{fig:SM_lipid_euclidean_vs_manifold_grid}
\end{figure}

We attempted to alter the kNN graph by adding a penalty $w_{i,j}=50\cdot w_{i, j}$ when two voxels belong to different anatomical regions according to the Allen annotation dataset. Figure \ref{fig:SM_lipid_euclidean_vs_manifold_grid} represents an illustration of the Implicit Manifold GP's ability to respect the anatomical boundary in the presence of the cross-boundary penalty, but we also can see that the underlying lipid measurements are not necessarily correlated with their proximity to the Allen boundaries.

We speculate that it's possible to further improve the results provided expert-constructed annotations specific to the lipid profiles in the MALDI dataset.

\FloatBarrier
\subsection{Latent Gaussian Process vs independent per-lipid Gaussian Process}

\begin{figure}[ht]
    \centering
    \includegraphics[width=0.5\linewidth]{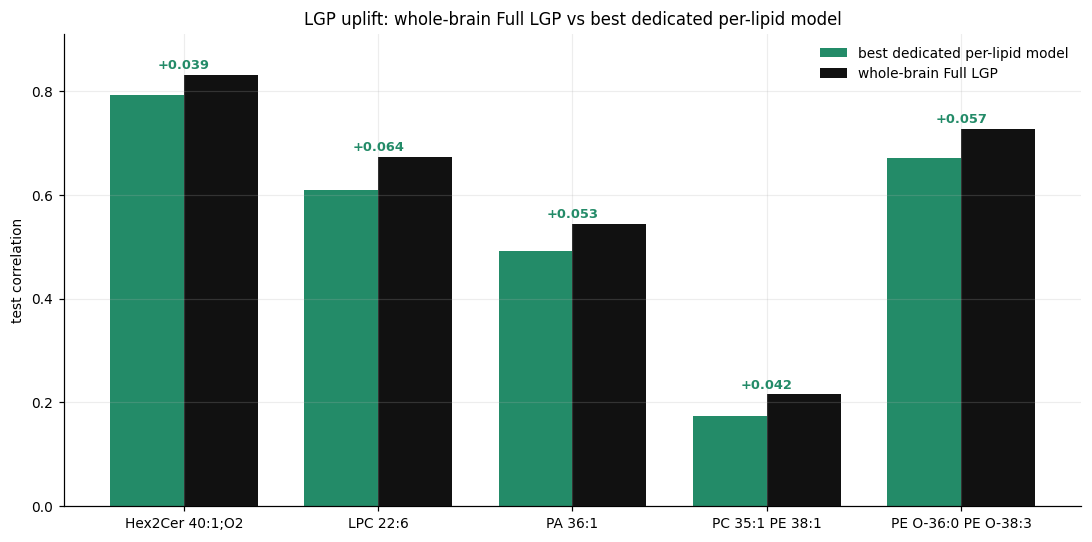}
    \caption{Per-lipid vs LGP uplift}
    \label{fig:per_lipid_vs_lgp_lift}
\end{figure}

Evaluating an independent GP per individual lipid is theoretically possible but turned out to be computationally infeasible and even undesirable given the uplift of adding an MLP in the latent space as seen in Figure \ref{fig:per_lipid_vs_lgp_lift}.

Using a limited number of GPs in a latent space potentially washes out the high frequency details of the individual lipids when these measurements are projected into the 5-dimensional space, but as seen in \ref{sec:maldi_analysis}, we can exploit the correlations between the lipids themselves to get a correlation uplift in the range $0.039 - 0.057$.

\section{Conclusions}

We showed that a simple LGP + MLP architecture (Fig \ref{fig:architecture}) provides a strong spatial reconstruction performance across sparsely sampled lipid sections of mouse brains, producing spatially  smooth reconstructions (\ref{fig:full_scale_render}), and achieving high correlation with measured lipid profiles. 

We compared the LGP architecture with a broad range of baselines, from linear interpolation, the original 3D interpolation from \cite{Bassini2025.10.13.682018} and standard MLPs to  recent state-of-the-art spatial-omics architectures, including NTF \cite{gong2026ultra}, SPA3D \cite{tang20263d}, and  DEEPSPATIAL \cite{yang2026reconstructing}. Our results demonstrate that the sparsity of the data set limits the ability of these models to exploit their underlying assumptions. In fact, we showed that even a simple Multi-Layer Perceptron can achieve higher correlation  than the recent state of the art spatial-omics models in this setting. We also evaluated GPLFR \cite{stevenson2026gaussian}, using  linear coregionalization instead of an MLP decoder and showed that nonlinear MLP decoding is better suited to capture relationships between measured profiles and latents.  

Finally, replacing Euclidean Mat\'ern Kernel in LGP with  an Implicit Manifold Gaussian Process \cite{fichera2023implicit} did not further improve reconstruction performance. The high voxel density of the reference model, together with  the lack of anatomical annotations specifically relevant for the lipid profiles, limits the benefits of the manifold-based approach.

\section*{Code availability}

The code is available at \url{https://github.com/SwissDataScienceCenter/mlibra}.

\section*{Acknowledgements}

This work was supported by the Swiss Data Science Center under grant agreement C22-07.

\bibliographystyle{ieeetr}
\bibliography{references}

\begin{thebibliography}{10}

\bibitem{Bassini2025.10.13.682018}
L.~F. Bassini, H.~H. Schede, L.~Capolupo, L.~H.~A. Alieh, F.~Venturi,
  A.~Valente, C.~Droin, D.~T. Banos, I.~Khven, E.~Z. Asirim, A.~Nashrallah,
  I.~Kaysudu, E.~Krymova, G.~D{\textquoteright}Angelo, and G.~L. Manno, ``The
  lipidomic architecture of the mouse brain,'' {\em bioRxiv}, 2025.

\bibitem{norris2013analysis}
J.~L. Norris and R.~M. Caprioli, ``Analysis of tissue specimens by
  matrix-assisted laser desorption/ionization imaging mass spectrometry in
  biological and clinical research,'' {\em Chemical reviews}, vol.~113, no.~4,
  pp.~2309--2342, 2013.

\bibitem{allen}
Q.~Wang, S.-L. Ding, Y.~Li, J.~Royall, D.~Feng, P.~Lesnar, N.~Graddis,
  M.~Naeemi, B.~Facer, A.~Ho, T.~Dolbeare, B.~Blanchard, N.~Dee, W.~Wakeman,
  K.~E. Hirokawa, A.~Szafer, S.~M. Sunkin, S.~W. Oh, A.~Bernard, J.~W.
  Phillips, M.~Hawrylycz, C.~Koch, H.~Zeng, J.~A. Harris, and L.~Ng, ``The
  allen mouse brain common coordinate framework: A 3d reference atlas,'' {\em
  Cell}, vol.~181, no.~4, pp.~936--953.e20, 2020.

\bibitem{fichera2023implicit}
B.~Fichera, S.~Borovitskiy, A.~Krause, and A.~G. Billard, ``Implicit manifold
  gaussian process regression,'' {\em Advances in neural information processing
  systems}, vol.~36, pp.~67701--67720, 2023.

\bibitem{stevenson2026gaussian}
E.~T. Stevenson, E.~T. Wolf, M.~T. Mak, N.~Mayne, and M.~Cranmer, ``Gaussian
  process latent factor regression for low-data, high-dimensional output
  problems,'' {\em arXiv preprint arXiv:2606.06576}, 2026.

\bibitem{zeira2022alignment}
R.~Zeira, M.~Land, A.~Strzalkowski, and B.~J. Raphael, ``Alignment and
  integration of spatial transcriptomics data,'' {\em Nature Methods}, vol.~19,
  no.~5, pp.~567--575, 2022.

\bibitem{hu2021spagcn}
J.~Hu, X.~Li, K.~Coleman, A.~Schroeder, N.~Ma, D.~J. Irwin, E.~B. Lee, R.~T.
  Shinohara, and M.~Li, ``Spagcn: Integrating gene expression, spatial location
  and histology to identify spatial domains and spatially variable genes by
  graph convolutional network,'' {\em Nature methods}, vol.~18, no.~11,
  pp.~1342--1351, 2021.

\bibitem{long2023spatially}
Y.~Long, K.~S. Ang, M.~Li, K.~L.~K. Chong, R.~Sethi, C.~Zhong, H.~Xu, Z.~Ong,
  K.~Sachaphibulkij, A.~Chen, {\em et~al.}, ``Spatially informed clustering,
  integration, and deconvolution of spatial transcriptomics with graphst,''
  {\em Nature communications}, vol.~14, no.~1, p.~1155, 2023.

\bibitem{zhao2021spatial}
E.~Zhao, M.~R. Stone, X.~Ren, J.~Guenthoer, K.~S. Smythe, T.~Pulliam, S.~R.
  Williams, C.~R. Uytingco, S.~E. Taylor, P.~Nghiem, {\em et~al.}, ``Spatial
  transcriptomics at subspot resolution with bayesspace,'' {\em Nature
  biotechnology}, vol.~39, no.~11, pp.~1375--1384, 2021.

\bibitem{jones2023alignment}
A.~Jones, F.~W. Townes, D.~Li, and B.~E. Engelhardt, ``Alignment of spatial
  genomics data using deep gaussian processes,'' {\em Nature methods}, vol.~20,
  no.~9, pp.~1379--1387, 2023.

\bibitem{tang20263d}
C.~Tang, Y.~Zhou, X.~Xiao, L.~Dong, L.~Yu, Q.~Li, G.~Xiao, and L.~Xu, ``3d
  reconstruction of spatial transcriptomics with spatial pattern enhanced graph
  convolutional neural network,'' {\em Briefings in bioinformatics}, vol.~27,
  no.~1, p.~bbag060, 2026.

\bibitem{gong2026ultra}
Y.~Gong, X.~Yuan, R.~Gao, J.~Chen, and Z.~Yu, ``Ultra-efficient high resolution
  3d reconstruction of spatial omics data with neural transcriptomic field,''
  {\em bioRxiv}, pp.~2026--05, 2026.

\bibitem{yang2026reconstructing}
Y.~Yang, Y.~Luo, K.~Zhang, Y.~Bu, Z.~Xia, H.~Peng, R.~Yan, Q.~Liu, D.~Lian,
  Y.~Chen, {\em et~al.}, ``Reconstructing true 3d spatial omics at single-cell
  resolution,'' {\em bioRxiv}, pp.~2026--04, 2026.

\bibitem{zhang2023molecularly}
M.~Zhang, X.~Pan, W.~Jung, A.~R. Halpern, S.~W. Eichhorn, Z.~Lei, L.~Cohen,
  K.~A. Smith, B.~Tasic, Z.~Yao, {\em et~al.}, ``Molecularly defined and
  spatially resolved cell atlas of the whole mouse brain,'' {\em Nature},
  vol.~624, no.~7991, pp.~343--354, 2023.

\bibitem{alvarez2012kernels}
M.~A. Alvarez, L.~Rosasco, and N.~D. Lawrence, ``Kernels for vector-valued
  functions: A review,'' {\em Foundations and Trends{\textregistered} in
  Machine Learning}, vol.~4, no.~3, pp.~195--266, 2012.

\bibitem{dai2017efficient}
Z.~Dai, M.~{\'A}lvarez, and N.~Lawrence, ``Efficient modeling of latent
  information in supervised learning using gaussian processes,'' {\em Advances
  in Neural Information Processing Systems}, vol.~30, 2017.

\bibitem{hensman2013gaussian}
J.~Hensman, N.~Fusi, and N.~D. Lawrence, ``Gaussian processes for big data,''
  {\em arXiv preprint arXiv:1309.6835}, 2013.

\bibitem{bruinsma2020scalable}
W.~Bruinsma, E.~Perim, W.~Tebbutt, S.~Hosking, A.~Solin, and R.~Turner,
  ``Scalable exact inference in multi-output gaussian processes,'' in {\em
  International Conference on Machine Learning}, pp.~1190--1201, PMLR, 2020.

\bibitem{higdon2008computer}
D.~Higdon, J.~Gattiker, B.~Williams, and M.~Rightley, ``Computer model
  calibration using high-dimensional output,'' {\em Journal of the American
  Statistical Association}, vol.~103, no.~482, pp.~570--583, 2008.

\bibitem{calandra2016manifold}
R.~Calandra, J.~Peters, C.~E. Rasmussen, and M.~P. Deisenroth, ``Manifold
  gaussian processes for regression,'' in {\em 2016 International joint
  conference on neural networks (IJCNN)}, pp.~3338--3345, IEEE, 2016.

\bibitem{rasmussen2006gpml}
C.~E. Rasmussen and C.~K.~I. Williams, {\em Gaussian Processes for Machine
  Learning}.
\newblock MIT Press, 2006.

\bibitem{whittle1963stochastic}
P.~Whittle, ``Stochastic processes in several dimensions,'' {\em Bulletin of
  the International Statistical Institute}, vol.~40, no.~2, pp.~974--994, 1963.

\bibitem{lindgren2011explicit}
F.~Lindgren, H.~Rue, and J.~Lindstr{\"o}m, ``An explicit link between gaussian
  fields and gaussian markov random fields: The stochastic partial differential
  equation approach,'' {\em Journal of the Royal Statistical Society: Series
  B}, vol.~73, no.~4, pp.~423--498, 2011.

\bibitem{feragen2015geodesic}
A.~Feragen, F.~Lauze, and S.~Hauberg, ``Geodesic exponential kernels: When
  curvature and linearity conflict,'' in {\em Proceedings of the IEEE
  conference on computer vision and pattern recognition}, pp.~3032--3042, 2015.

\bibitem{belkin2003laplacian}
M.~Belkin and P.~Niyogi, ``Laplacian eigenmaps for dimensionality reduction and
  data representation,'' {\em Neural Computation}, vol.~15, no.~6,
  pp.~1373--1396, 2003.

\bibitem{coifman2006diffusion}
R.~R. Coifman and S.~Lafon, ``Diffusion maps,'' {\em Applied and Computational
  Harmonic Analysis}, vol.~21, no.~1, pp.~5--30, 2006.
\newblock Introduces the alpha-family of density normalizations; alpha=1
  recovers the Laplace--Beltrami operator.

\bibitem{borovitskiy2021graph}
V.~Borovitskiy, I.~Azangulov, A.~Terenin, P.~Mostowsky, M.~P. Deisenroth, and
  N.~Durrande, ``Mat\'ern gaussian processes on graphs,'' in {\em Artificial
  Intelligence and Statistics (AISTATS)}, 2021.

\bibitem{hein2005graphs}
M.~Hein, J.-Y. Audibert, and U.~von Luxburg, ``From graphs to manifolds---weak
  and strong pointwise consistency of graph laplacians,'' in {\em Conference on
  Learning Theory (COLT)}, 2005.

\bibitem{vonluxburg2007tutorial}
U.~von Luxburg, ``A tutorial on spectral clustering,'' {\em Statistics and
  Computing}, vol.~17, no.~4, pp.~395--416, 2007.

\bibitem{williams2001nystrom}
C.~K.~I. Williams and M.~Seeger, ``Using the nystr{\"o}m method to speed up
  kernel machines,'' in {\em Advances in Neural Information Processing Systems
  (NIPS)}, 2001.

\bibitem{bengio2004out}
Y.~Bengio, J.-F. Paiement, P.~Vincent, O.~Delalleau, N.~Le~Roux, and M.~Ouimet,
  ``Out-of-sample extensions for lle, isomap, mds, eigenmaps, and spectral
  clustering,'' in {\em Advances in Neural Information Processing Systems
  (NIPS)}, 2004.

\bibitem{titsias2009variational}
M.~K. Titsias, ``Variational learning of inducing variables in sparse gaussian
  processes,'' in {\em Artificial Intelligence and Statistics (AISTATS)}, 2009.

\end{thebibliography}

\appendix
\setcounter{figure}{0}
\setcounter{table}{0}
\renewcommand{\thefigure}{A\arabic{figure}}
\renewcommand{\thetable}{A\arabic{table}}

\section{Appendix: Hyperparameters}
\label{sec:hyperparameters}

\begin{table}[h]
\centering
\caption{Hyperparameters swept in the experiments. Defaults are the values used
in the reported production runs; ``--'' denotes a fixed (unswept) setting listed
for completeness.}
\label{tab:hyperparams}
\small
\begin{tabular}{lllc}
\toprule
Hyperparameter & Symbol & Swept values & Default \\
\midrule
\multicolumn{4}{l}{\textit{Graph construction}}\\
$k$-NN neighbours          & $k$          & $10,\,15,\,30,\,60,\,120$                 & $15$ \\
Graph bandwidth            & $\varepsilon$& $0.01,\,0.02,\,0.05,\,0.1,\,0.2,\,0.5,\,1.0$ & $0.1$ \\
Laplacian normalization    & --           & symmetric, random-walk                    & random-walk \\
Node stride                & --           & $4,\,8$                                   & $4$ \\
Tissue threshold           & --           & $5,\,40,\,50$                             & $5$ \\
\addlinespace
\multicolumn{4}{l}{\textit{Anatomical edge handling}}\\
$k$-NN weighting           & --           & faiss, atlas-weighted, cluster-weighted, anatomical & atlas-weighted \\
Cross-region inflation     & $\gamma$     & $10,\,50$                                 & $50$ \\
Root handling              & --           & dissolve, ignore, cross                   & dissolve \\
Denoise (majority-vote its)& --           & $0,\,3$                                   & $3$ \\
Prune cross-region frac.   & --           & $0,\,0.95,\,0.97$                         & $0.97$ \\
\addlinespace
\multicolumn{4}{l}{\textit{Kernel spectrum}}\\
Number of modes            & $K$          & $100,\,300,\,1000,\,1300,\,2300$          & $300$ / $1300$\textsuperscript{a} \\
Mat\'ern smoothness        & $\nu$        & (fixed)                                    & $2$ \\
Product-ARD factor         & --           & on, off                                    & off \\
\addlinespace
\multicolumn{4}{l}{\textit{Nystr\"om out-of-sample (bump)}}\\
Bump scale                 & --           & $1,\,2,\,3,\,5,\,10,\,20$                 & $1$ \\
Bump decay                 & --           & $0.01,\,0.05,\,0.1$                       & $0.01$ \\
\addlinespace
\multicolumn{4}{l}{\textit{\textsc{euclid} interpolation baseline}\textsuperscript{b}}\\
Intensity filter           & $w$ & $0,\,50$                    & $0$ \\
Intensity normalization    & --  & none, per-lipid max         & none\textsuperscript{c} \\
Structure gate (annotation)& --  & Allen leaf (672 lab.), \texttt{level\_15annot} (70 lab.) & \texttt{level\_15annot} \\
Donor sphere radius        & --  & --  & $26$ vox ($2.6$\,mm) \\
Distance kernel            & --  & --  & $\exp(-d)$, $d$ in vox \\
Interpolation grid         & --  & --  & $100\,\mu$m CCF \\
Unfilled-voxel backfill    & --  & --  & $10^3$ box mean \\
\addlinespace
\multicolumn{4}{l}{\textit{Model / training (fixed)}}\\
Latent dimension           & $d$          & --                                        & $5$ \\
Inducing points            & $M$          & --                                        & $1000$ \\
Batch size / epochs / lr   & --           & --                                        & $1000$ / $30$ / $10^{-3}$ \\
\addlinespace
\multicolumn{4}{l}{\textit{Evaluation}}\\
Cross-validation           & --           & 8 folds, leave-one-out, leave-one-section-out & fold~2 \\
\bottomrule
\end{tabular}
\end{table}

\begin{table}[t]
\centering
\caption{Baseline and state-of-the-art comparison models with their principal
hyperparameters. Decoder/GCN widths are matched to our models where they are
shared plumbing; method-defining settings follow each paper.}
\label{tab:models}
\small
\begin{tabular}{lll}
\toprule
Model & Mechanism & Key hyperparameters \\
\midrule
\multicolumn{3}{l}{\textit{Baselines}}\\
Mean        & per-lipid training mean                 & --- \\
Linear      & ridge regression (closed form)          & $\alpha=1$ \\
XGBoost     & $P$ boosted-tree regressors             & $400$ trees, depth $6$, lr $0.05$, early stop \\
MLP         & coordinate$\rightarrow$lipid MLP        & $256\text{-}256\text{-}128$, SiLU, dropout $0.1$ \\
MLP-bottleneck & MLP with width-5 bottleneck          & $256\text{-}5\text{-}256\text{-}256\text{-}128$ \\
MLP-eigen   & MLP on Laplacian eigenmap features      & $256\text{-}256\text{-}128$; $K$ eigenmodes \\
EUCLID \cite{Bassini2025.10.13.682018} & structure-gated $\exp(-d)$ interpolation & radius $26$ vox ($2.6$\,mm), same-leaf-label \\
            &                                         & donors, $100\,\mu$m CCF grid, $w{=}0$, $10^3$ box backfill \\
GCN         & graph conv.\ net over $k$-NN coords     & $512\text{-}512\text{-}256$, dropout $0.1$, $k{=}15$, interp $8$ \\
\addlinespace
\multicolumn{3}{l}{\textit{State of the art}}\\
NTF \cite{gong2026ultra}    & hash-grid implicit neural field & $12$ levels $\times 8$ feat., hashmap $2^{19}$, res $16{\to}1024$, \\
            &                                         & MLP $256\text{-}256\text{-}128$, TV $0.05$, PSF $4/\sigma0.01$ \\
SPA3D \cite{tang20263d}& SPE denoise + $z$-aware GCN      & ALFT keep $0.5$, grid $128$, $z$-weight $0.3$, \\
            &                                         & GCN $512\text{-}512\text{-}256$, $k{=}15$, $8{\times}10^4$ nodes \\
DEEPSPATIAL \cite{yang2026reconstructing} & flow matching + UOT & hidden $256$, depth $6$, $8$ heads, patch $8$, \\
            &                                         & ODE $50$ steps, UOT reg $0.3$/$\tau0.05$, cross-mouse \\
GPLFR \cite{stevenson2026gaussian} & latent GP + marg.\ linear decoder & $d{=}8$, $256$ inducing, Mat\'ern $\nu{=}5/2$, base=Riemann \\
\addlinespace
\multicolumn{3}{l}{\textit{Proposed}}\\
Euclidean SVGP     & inducing SVGP + MLP decoder        & $1000$ inducing, Mat\'ern, $d{=}5$ \\
Manifold GP (ind.) & graph-Laplacian kernel, inducing   & $K{=}300$, $1000$ inducing, $d{=}5$ \\
Spectral Manifold  & weight-space (diagonal $q(\mathbf w)$) & $K{=}1300$, $d{=}5$ \\
\bottomrule
\end{tabular}
\end{table}

\begin{table}[htbp]
  \centering
  \small
  \makebox[\textwidth][c]{\begin{tabular}{llp{0.52\linewidth}}
    \toprule
    Token & Parameter & Meaning \\
    \midrule
    \multicolumn{3}{l}{\emph{Graph \& spectral basis}}\\
    \texttt{K$n$} & \texttt{NUM\_MODES} & Laplacian eigenmodes retained in the kernel basis \\
    \texttt{faiss} & \texttt{KNN\_METHOD} & plain $k$NN graph, unweighted by atlas \\
    \texttt{faiss\_atlas\_weighted} & \texttt{KNN\_METHOD} & cross-region edges inflated using Allen atlas labels \\
    \texttt{faiss\_cluster\_weighted} & \texttt{KNN\_METHOD} & as above, but regions from data-driven clustering \\
    \texttt{x50} & \texttt{CROSS\_REGION\_INFLATION} & multiplier on $d^2$ for cross-region edges \\
    \texttt{randomwalk} & \texttt{LAPLACIAN\_NORM} & $L_{\mathrm{rw}} = I - D^{-1}W$ (vs.\ \texttt{symmetric}) \\
    \texttt{prune$p$} & \texttt{PRUNE\_CROSS\_REGION} & drop cross-region edges above quantile $p$ \\
    \texttt{dn$n$} & \texttt{DENOISE\_LABELS} & majority-vote label smoothing, radius $n$ \\
    \texttt{rootdissolve} & \texttt{ROOT\_HANDLING} & dissolve the atlas root (997) catch-all label \\
    \texttt{level\_$n$annot} & atlas file & Allen annotation level ($15$ = default) \\
    \midrule
    \multicolumn{3}{l}{\emph{Kernel}}\\
    \texttt{learndiff$v$} & \texttt{DIFFUSION\_SCALE} & learnable diffusion scale, init $v$ (\texttt{diff}\,=\,frozen) \\
    \texttt{ard} & \texttt{--ard} & per-axis ARD lengthscales instead of isotropic \\
    \texttt{prodard$\nu$} & \texttt{PRODUCT\_ARD\_NU} & product-ARD kernel with smoothness $\nu$ \\
    \midrule
    \multicolumn{3}{l}{\emph{Inducing points \& sampling}}\\
    \texttt{M$n$} & \texttt{NUM\_INDUCING\_POINTS} & number of inducing points \\
    \texttt{learnind} & \texttt{--learn-inducing} & inducing locations optimised, not fixed \\
    \texttt{data} / \texttt{reference} & \texttt{INDUCING\_SOURCE} & inducing points drawn from data vs.\ reference template \\
    \texttt{blend$f$} & \texttt{INDUCING\_DENSITY\_FRAC} & fraction of inducing points drawn density-proportionally \\
    \texttt{RSAMPLE} / \texttt{MEAN} & \texttt{NO\_RSAMPLE} & reparameterised sample vs.\ posterior mean during training \\
    \midrule
    \multicolumn{3}{l}{\emph{\textsc{euclid} interpolation}\textsuperscript{$\dagger$}}\\
    \texttt{w$v$}      & \texttt{EUCLID\_W}     & intensity filter on EUCLID's $0$--$255$ rescale: voxels below $v$ stop being donors and are re-imputed ($v{=}50$ is their rendering default; $0$ disables the filter) \\
    \texttt{norm$m$}   & \texttt{EUCLID\_NORM}  & intensity normalisation before that rescale: \texttt{max} = per-lipid training max (\texttt{none} = as-is) \\
    \texttt{euclidatlas} & \texttt{EUCLID\_ATLAS} & EUCLID's shipped $100\,\mu$m Allen leaf annotation ($672$ labels) as the same-structure donor gate \\
    \texttt{level\_$n$annot-reference\_image} & \texttt{EUCLID\_ATLAS} & our annotation/reference pair as the gate instead, subsampled to $100\,\mu$m \\
    
    \midrule
    \multicolumn{3}{l}{\emph{Baselines \& SOTA}}\\
    \texttt{clk$k$}, \texttt{sw$w$}, \texttt{cs$s$} & clustering & cluster count, spatial weight, seed (spectral GP) \\
    \texttt{NM$n$} & modes & eigenmodes fed to the MLP-eigen baseline \\
    \texttt{ZW$z$} & \texttt{--spa3d-z-weight} & anisotropic scaling of the inter-slice ($z$) axis \\
    \texttt{SPE} & \texttt{--spa3d-spe} & spatial pattern enhancement: \texttt{none} / \texttt{alft} / \texttt{hilbert} \\
    \texttt{RES$r$} & \texttt{--ntf-max-res} & max resolution of the multi-resolution hash grid \\
    \texttt{TV$w$} & \texttt{--ntf-tv-weight} & total-variation smoothness weight \\
    \texttt{d$n$} & latent dim & number of latent factors (GPLFR) \\
    \texttt{bs$n$} & batch size & minibatch size (baselines) \\
    \bottomrule
  \end{tabular}
  }
  \caption{Legend for the configuration tokens in Table~\ref{tab:winning-configs}.}
  \label{tab:config-legend}
\end{table}

\section{Appendix: Rendered images for two lipids across all models}
\label{sec:renders}

\begin{figure}[htbp]
    \centering
    \includegraphics[width=0.9\linewidth]{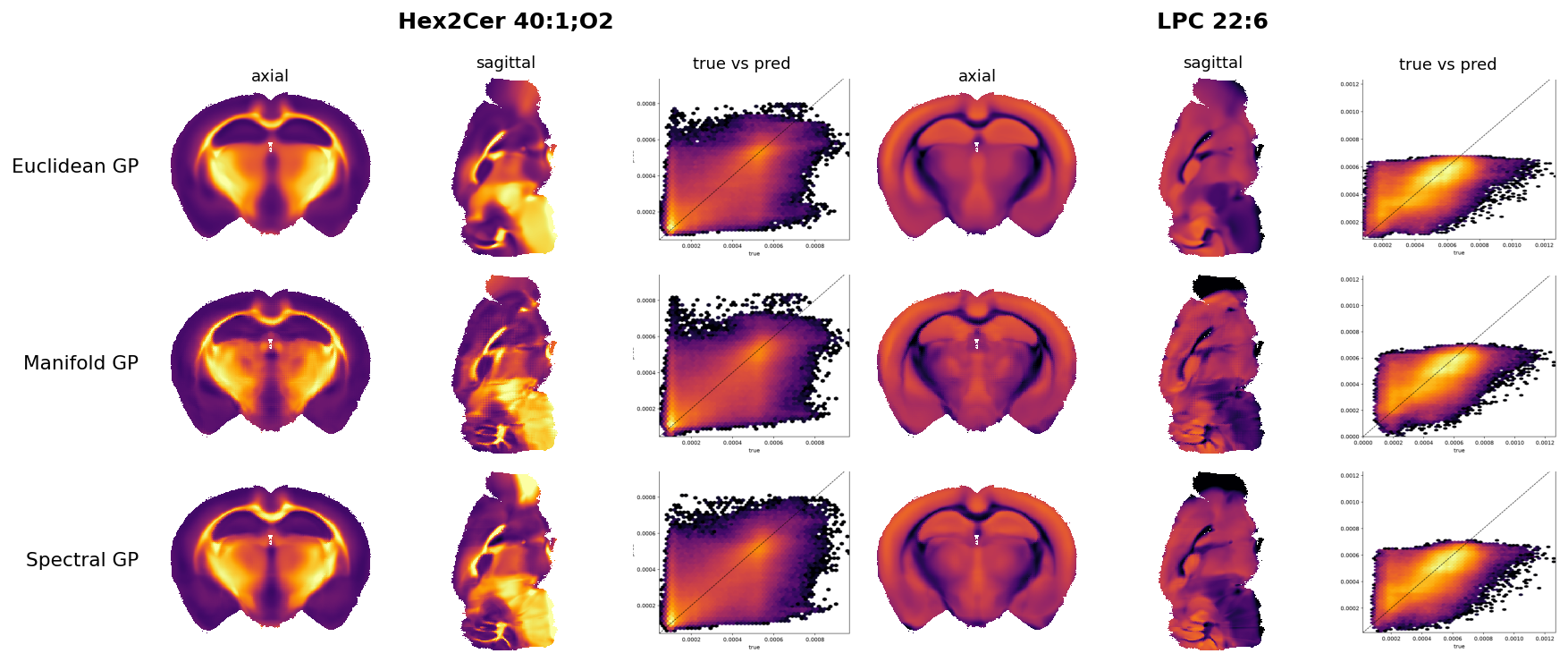}
    \caption{Example lipid profile generations using Latent Gaussian Process models}
    \label{fig:combined_gps}
\end{figure}

\begin{figure}[htbp]
    \centering
    \includegraphics[width=0.9\linewidth]{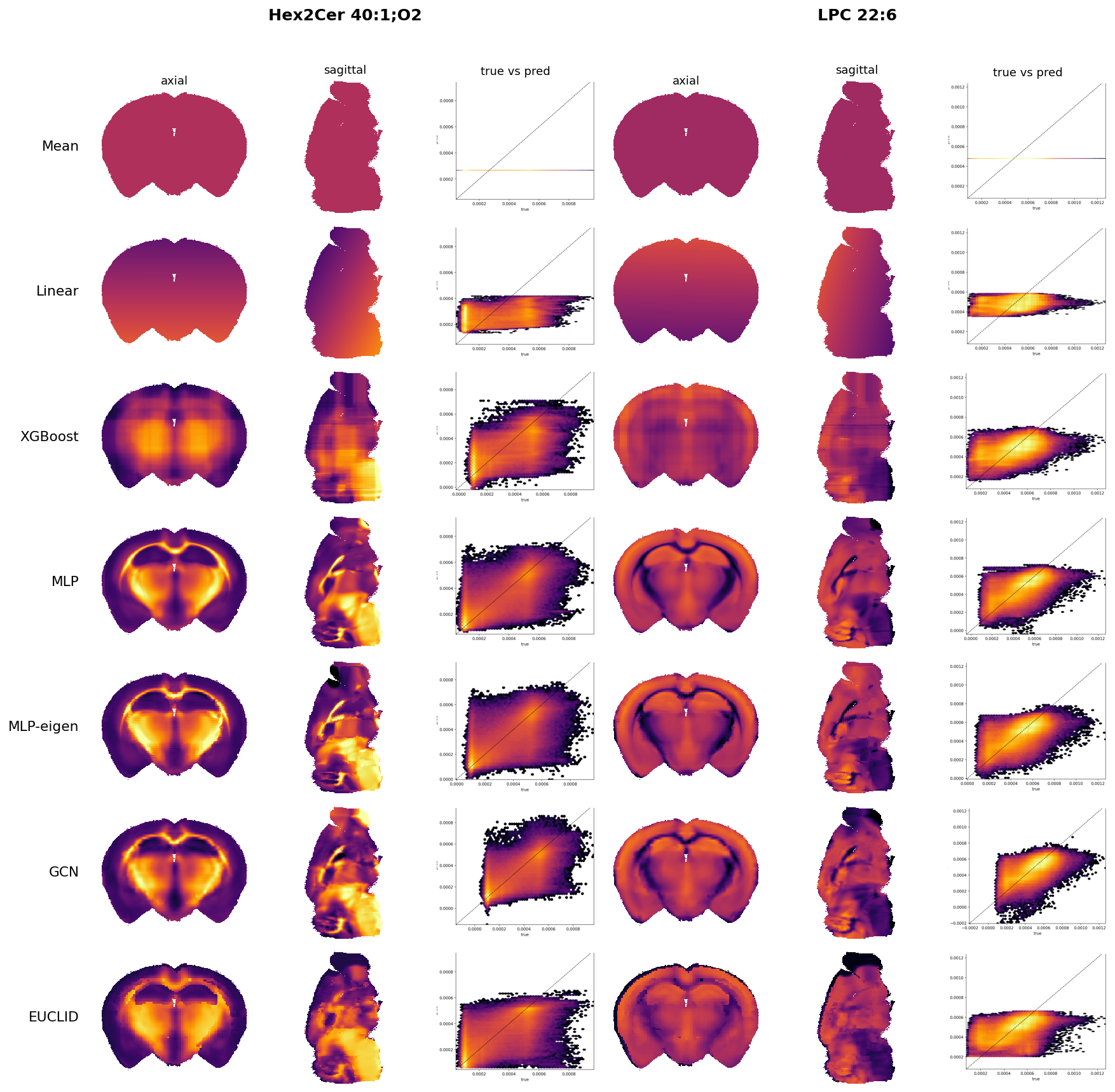}
    \caption{Example lipid profile generations using baseline models}
    \label{fig:combind_baselines}
\end{figure}

\begin{figure}[htbp]
    \centering
    \includegraphics[width=0.9\linewidth]{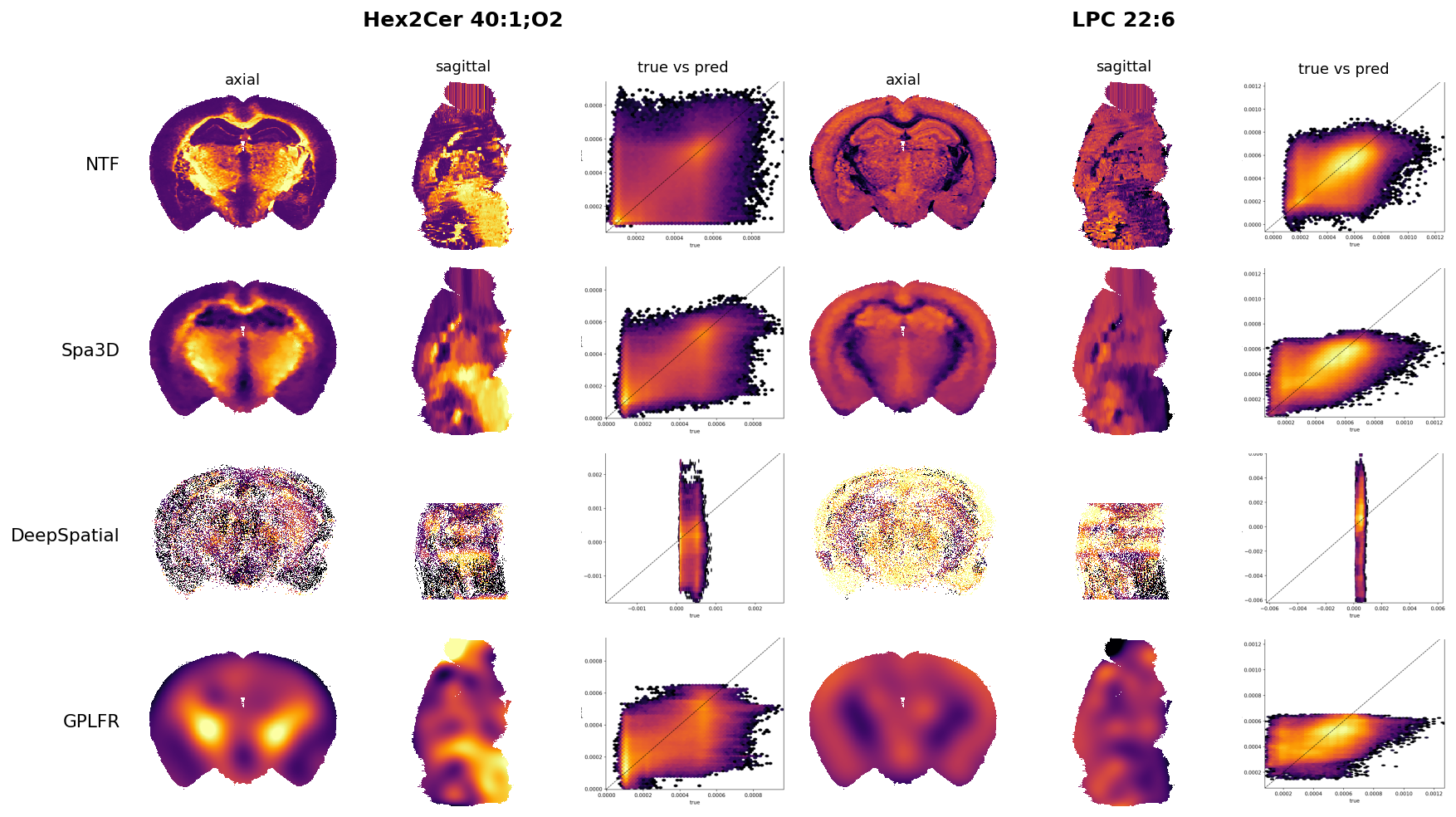}
    \caption{Example lipid profile generations using State of the Art models}
    \label{fig:combind_sota}
\end{figure}

\section{Mathematical background}
\label{sec:math_background}

\subsection{Gaussian processes}
\label{sec:app_gp}

A Gaussian process generalizes the multivariate Gaussian distribution to
infinite dimensions: instead of sampling vectors, it samples \emph{functions}
\cite{rasmussen2006gpml}. A GP is fully specified by a mean function
$m(\mathbf x)=\mathbb E[f(\mathbf x)]$---almost always taken as $m\equiv 0$ after
centering, or as a linear mean as in our model---and a covariance (kernel)
function $k(\mathbf x,\mathbf x')=\mathbb E[(f(\mathbf x)-m(\mathbf x))(f(\mathbf
x')-m(\mathbf x'))]$, and is defined by the property that any finite collection
of function values is jointly Gaussian,
\begin{equation}
f\sim\mathcal{GP}\big(m(\mathbf x),\,k(\mathbf x,\mathbf x')\big)
\quad\Longleftrightarrow\quad
\mathbf f(X)=\big(f(\mathbf x_1),\dots,f(\mathbf x_N)\big)^\top
\sim\mathcal N\big(m(X),\,K(X,X)\big)
\end{equation}
for every finite $X=\{\mathbf x_i\}_{i=1}^N$. The kernel must therefore be
symmetric and positive semi-definite (PSD): every Gram matrix
$K_{ij}=k(\mathbf x_i,\mathbf x_j)$ it produces must be a valid covariance
matrix. This requirement is not a technicality---it is precisely what fails for
naive geodesic kernels in Sec.~\ref{sec:app_manifold}.

Given training data $\mathcal D=\{(\mathbf x_i,y_i)\}_{i=1}^N$ with i.i.d.
Gaussian observation noise,
\begin{equation}
\mathbf y=\mathbf f(X)+\boldsymbol\varepsilon,
\qquad \boldsymbol\varepsilon\sim\mathcal N(\mathbf 0,\sigma_n^2\mathbf I),
\end{equation}
and test inputs $X_*$, we write $K=K(X,X)\in\mathbb R^{N\times N}$,
$K_*=K(X,X_*)\in\mathbb R^{N\times N_*}$ and
$K_{**}=K(X_*,X_*)\in\mathbb R^{N_*\times N_*}$. Conditioning the joint Gaussian
on $\mathbf y$ gives the posterior
\begin{equation}
\mathbf f_*\mid X,\mathbf y,X_*\sim\mathcal N(\boldsymbol\mu_*,\Sigma_*),
\qquad
\boldsymbol\mu_*=K_*^\top\big[K+\sigma_n^2\mathbf I\big]^{-1}\mathbf y,
\qquad
\Sigma_*=K_{**}-K_*^\top\big[K+\sigma_n^2\mathbf I\big]^{-1}K_*,
\label{eq:gp_posterior}
\end{equation}
where $\boldsymbol\mu_*$ is the prediction and $\Sigma_*$ the calibrated
uncertainty. Two properties matter for our application: the posterior mean is a
smooth interpolant that behaves well with few observations, and the posterior
variance in \eqref{eq:gp_posterior} is available at every location and does not
depend on the observed values---only on where we sampled.

Kernel hyperparameters $\boldsymbol\theta$ (lengthscale, output scale, noise) are
learned by maximizing the log marginal likelihood,
\begin{equation}
\log p(\mathbf y\mid X,\boldsymbol\theta)
=-\tfrac12\mathbf y^\top\big[K_{\boldsymbol\theta}+\sigma_n^2\mathbf I\big]^{-1}\mathbf y
-\tfrac12\log\big|K_{\boldsymbol\theta}+\sigma_n^2\mathbf I\big|
-\tfrac N2\log 2\pi ,
\label{eq:lml}
\end{equation}
whose two data-dependent terms trade data fit against model complexity. Both
\eqref{eq:gp_posterior} and \eqref{eq:lml} require a Cholesky factorization of an
$N\times N$ matrix: $\mathcal O(N^3)$ time and $\mathcal O(N^2)$ memory. With
$N\sim10^5$ pixels this is infeasible, which motivates the use of Stochastic Variational Gaussian Processes
Sec.~\ref{sec:app_svgp}.

\subsection{Stationary kernels}
\label{sec:app_kernels}

The kernels used in practice are almost always \emph{stationary}: they depend on
the inputs only through the Euclidean displacement $\mathbf x-\mathbf x'$, and
usually only through the distance $r=\|\mathbf x-\mathbf x'\|$. The two canonical
choices are the squared exponential (RBF) kernel and the Mat\'ern family:
\begin{equation}
k_{\mathrm{RBF}}(\mathbf x,\mathbf x')=\sigma^2\exp\!\Big(-\frac{r^2}{2\ell^2}\Big),
\qquad
k_{\nu}(\mathbf x,\mathbf x')=\sigma^{2}\,\frac{2^{1-\nu}}{\Gamma(\nu)}
\Big(\frac{\sqrt{2\nu}\,r}{\ell}\Big)^{\nu}
\mathcal K_{\nu}\!\Big(\frac{\sqrt{2\nu}\,r}{\ell}\Big),
\label{eq:matern}
\end{equation}
with $\mathcal K_\nu$ the modified Bessel function of the second kind. The
smoothness parameter $\nu$ controls mean-square differentiability: samples are
$\lceil\nu\rceil-1$ times differentiable, so $\nu=1/2$ gives the
(non-differentiable) exponential kernel, $\nu=3/2$ and $\nu=5/2$ the standard
rough/smooth compromises used in spatial modelling, and $\nu\to\infty$ recovers
$k_{\mathrm{RBF}}$. The lengthscale $\ell$ sets the distance over which the
function decorrelates and $\sigma^2$ the marginal variance. Both kernels are
isotropic; anisotropy is obtained by replacing $r^2$ with the automatic
relevance determination (ARD) form $\sum_{d}(x_d-x'_d)^2/\ell_d^2$, which lets
the model learn a separate scale per input dimension.

By Bochner's theorem a stationary kernel on $\mathbb R^d$ is PSD if and only if
it is the Fourier transform of a non-negative spectral measure. The Mat\'ern
kernel corresponds to the spectral density
\begin{equation}
S(\boldsymbol\omega)\ \propto\ \Big(\frac{2\nu}{\ell^{2}}+\|\boldsymbol\omega\|^{2}\Big)^{-\left(\nu+\frac d2\right)} ,
\label{eq:matern_spectral}
\end{equation}
equivalently it is the stationary solution of the Whittle stochastic PDE
\cite{whittle1963stochastic,lindgren2011explicit}
\begin{equation}
\Big(\frac{2\nu}{\ell^{2}}-\Delta\Big)^{\frac{\nu}{2}+\frac d4} f=\mathcal W ,
\label{eq:spde}
\end{equation}
with $\Delta$ the Laplacian and $\mathcal W$ Gaussian white noise. Equation
\eqref{eq:spde} is the characterization that generalizes: it never mentions the
distance $r$, only a differential operator.

\subsection{From Euclidean to manifold kernels}
\label{sec:app_manifold}

Suppose the inputs lie on a $d$-dimensional Riemannian manifold
$\mathcal M$ embedded in $\mathbb R^D$---for us, the brain volume, whose
anatomy makes tissue-intrinsic proximity differ from ambient proximity. The
tempting move is to keep \eqref{eq:matern} and replace the Euclidean distance
$r$ by the geodesic distance $d_{\mathcal M}(\mathbf x,\mathbf x')$. This is
\emph{invalid}: \cite{feragen2015geodesic} show that the geodesic Gaussian
kernel $\exp(-d_{\mathcal M}^2/2\ell^2)$ is PSD for all $\ell>0$ only if the
metric space is flat, i.e. isometrically embeddable in a Hilbert space; on a
curved manifold there exist point configurations whose Gram matrix has negative
eigenvalues, so the ``covariance'' is not a covariance at all. The same
obstruction applies to the geodesic Mat\'ern form.

The resolution of \cite{fichera2023implicit} is to generalize the
\emph{characterization of the operator} \eqref{eq:spde} rather than the distance: keep
the SPDE exactly as it is and replace the Euclidean Laplacian $\Delta$ by the
Laplace--Beltrami operator $\Delta_{\mathcal M}$ of the manifold,
\begin{equation}
\Big(\frac{2\nu}{\ell^{2}}-\Delta_{\mathcal M}\Big)^{\frac{\nu}{2}+\frac d4} f=\mathcal W .
\label{eq:spde_manifold}
\end{equation}
On a compact manifold $\Delta_{\mathcal M}$ is positive
semi-definite with a discrete spectrum $\Delta_{\mathcal
M}\phi_k=\lambda_k\phi_k$, $0=\lambda_0\le\lambda_1\le\cdots$, whose
eigenfunctions form an orthonormal basis of $L^2(\mathcal M)$. Solving
\eqref{eq:spde_manifold} in that basis gives the Riemannian Mat\'ern kernel
\begin{equation}
k(\mathbf x,\mathbf x')=\frac{\sigma^2}{C}\sum_{k=0}^{\infty}
S(\lambda_k)\,\phi_k(\mathbf x)\,\phi_k(\mathbf x'),
\qquad
S(\lambda)=\Big(\frac{2\nu}{\ell^{2}}+\lambda\Big)^{-\left(\nu+\frac d2\right)},
\label{eq:riemannian_matern}
\end{equation}
with $C$ a normalizing constant fixing the marginal variance. This is PSD
\emph{by construction}: it is an explicit Mercer expansion with non-negative
coefficients, so every Gram matrix it produces is the Gram matrix of the feature
map $\Phi(\mathbf x)=\big(\sqrt{S(\lambda_k)}\,\phi_k(\mathbf x)\big)_k$. It
reduces to \eqref{eq:matern} when $\mathcal M=\mathbb R^d$, since
\eqref{eq:matern_spectral} is exactly $S$ evaluated on the Euclidean spectrum.

Two consequences deserve emphasis, because they are frequently misread. First,
the resulting kernel is \emph{not} a function of the geodesic distance. It is a
function of the whole eigenfunction profile at the two points---a \emph{spectral}
similarity---and geodesic proximity is encoded only implicitly, through the fact
that low-frequency eigenfunctions vary slowly along the manifold. Second, the
kernel is only \emph{isotropic in the spectrum}: the filter $S$ acts on
$\lambda$, not on position, so all spatial structure enters through the
eigenbasis.

\subsection{Graph discretization}
\label{sec:app_graph}

$\mathcal M$ and $\Delta_{\mathcal M}$ are unknown; we observe a point cloud
$\{\mathbf x_i\}_{i=1}^N$ sampled from it. Following the graph-Laplacian
literature \cite{belkin2003laplacian,coifman2006diffusion} and its
GP instantiation \cite{borovitskiy2021graph,fichera2023implicit}, we build a
symmetric $k$-nearest-neighbour graph on the point cloud with heat weights of
bandwidth $\epsilon$,
\begin{equation}
w_{ij}=\exp\!\Big(-\frac{\|\mathbf x_i-\mathbf x_j\|^{2}}{4\epsilon^{2}}\Big)
\quad\text{for } j\in\mathcal N_k(i)\ \text{(and symmetrized)},\qquad w_{ij}=0\ \text{otherwise},
\label{eq:heat_weights}
\end{equation}
so that only \emph{neighbouring} points interact directly through their
Euclidean distance---which on a manifold agrees with the geodesic distance to
first order---while distant points are related only through paths in the graph.
Because the sampling density of the point cloud is non-uniform, we apply the
\cite{coifman2006diffusion} $\alpha=1$ density normalization before forming the
Laplacian,
\begin{equation}
d_i=\sum_j w_{ij},\qquad
\widetilde w_{ij}=\frac{w_{ij}}{d_i\,d_j},\qquad
\widetilde d_i=\sum_j\widetilde w_{ij},
\label{eq:alpha1}
\end{equation}
which cancels the sampling density to first order, and take the random-walk
graph Laplacian
\begin{equation}
\boldsymbol\Delta_{\mathrm{rw}}
=\frac{1}{\epsilon^{2}}\big(\mathbf I-\widetilde{\mathbf D}^{-1}\widetilde{\mathbf W}\big)
=\widetilde{\mathbf D}^{-1/2}\boldsymbol\Delta_{\mathrm{sym}}\widetilde{\mathbf D}^{1/2},
\qquad
\boldsymbol\Delta_{\mathrm{sym}}
=\frac{1}{\epsilon^{2}}\big(\mathbf I-\widetilde{\mathbf D}^{-1/2}\widetilde{\mathbf W}\widetilde{\mathbf D}^{-1/2}\big).
\label{eq:laplacians}
\end{equation}
Its bottom $K$ eigenpairs $\{\lambda_k,\boldsymbol\phi_k\}_{k<K}$ are extracted, and \eqref{eq:riemannian_matern}
is truncated to those $K$ modes:
\begin{equation}
k(\mathbf x,\mathbf x')=\sum_{k=0}^{K-1}\widetilde S(\lambda_k)\,
\phi_k(\mathbf x)\,\phi_k(\mathbf x')
=\big\langle \Phi(\mathbf x),\Phi(\mathbf x')\big\rangle,
\qquad
\Phi_k(\mathbf x)=\sqrt{\widetilde S(\lambda_k)}\;\phi_k(\mathbf x),
\label{eq:manifold_kernel}
\end{equation}
where, following the graph Mat\'ern kernel of \cite{borovitskiy2021graph}, the
filter is
\begin{equation}
S(\lambda)=\Big(\frac{2\nu}{\ell^{2}}+\kappa\lambda\Big)^{-\nu},
\qquad
\widetilde S(\lambda_k)=N\,\frac{S(\lambda_k)}{\sum_{j<K}S(\lambda_j)} .
\label{eq:filter}
\end{equation}

Three properties of \eqref{eq:manifold_kernel} are worth stating explicitly.
(i) It is an inner product of explicit $K$-dimensional features, hence PSD, and
evaluating it costs $\mathcal O(K)$ per pair with no $N\times N$ matrix ever
formed. (ii) It has \emph{rank at most $K$}. Consequently any inducing set with
$M>K$ points produces a singular $K_{uu}$ in Sec.~\ref{sec:app_svgp} and adds no
model capacity---the mode budget, not the inducing budget, is the binding
constraint. (iii) The truncation is a low-pass approximation: the finest spatial
detail the kernel can express is set by $K$, while $K$ larger than the node
density supports merely adds aliased, noise-dominated eigenvectors.

\paragraph{Remark: only the random-walk normalization converges to
Laplace--Beltrami.} The three graph Laplacians in common use---unnormalized,
symmetric, and random-walk---are not interchangeable in
\eqref{eq:manifold_kernel}. Only $\boldsymbol\Delta_{\mathrm{rw}}$ is a pointwise
consistent estimator of $\Delta_{\mathcal M}$; the unnormalized and symmetric
versions converge to $\Delta_{\mathcal M}$ \emph{plus} a first-order drift term
involving the sampling density, i.e. they estimate a density-weighted operator
\cite{coifman2006diffusion,hein2005graphs,vonluxburg2007tutorial}. We therefore
solve the eigenproblem in the symmetric form, which is numerically preferable,
and recover the random-walk eigenfunctions by the similarity transform implied by
\eqref{eq:laplacians}, $\boldsymbol\phi_k^{\mathrm{rw}}=\widetilde{\mathbf
D}^{-1/2}\boldsymbol\phi_k^{\mathrm{sym}}$, the eigenvalues being unchanged.
These eigenvectors are orthonormal in the degree-weighted inner product
$\langle u,v\rangle_{\widetilde{\mathbf D}}=u^\top\widetilde{\mathbf D}v$ rather
than the Euclidean one, so any projection of a field onto the eigenbasis must use
$c_k=\boldsymbol\phi_k^\top\widetilde{\mathbf D}\,\mathbf f$.

\subsection{Out-of-sample extension}
\label{sec:app_nystrom}

Equation \eqref{eq:manifold_kernel} defines $\phi_k$ only at the $N$ graph nodes,
whereas prediction requires $\phi_k(\mathbf x_*)$ at arbitrary $\mathbf
x_*\notin\{\mathbf x_i\}$. We extend the eigenfunctions by the Nystr\"om method
\cite{williams2001nystrom,bengio2004out}: form the same heat weights and
$\alpha=1$ normalization \eqref{eq:heat_weights}--\eqref{eq:alpha1} between
$\mathbf x_*$ and its $k$ nearest graph nodes and row-normalize them into a
transition distribution $p(\mathbf x_*,\cdot)$, then average the neighbouring
eigenvector values. On the graph itself the transition matrix is
$\mathbf P=\widetilde{\mathbf D}^{-1}\widetilde{\mathbf W}=\mathbf
I-\epsilon^{2}\boldsymbol\Delta_{\mathrm{rw}}$, which acts on mode $k$ as
$\mathbf P\boldsymbol\phi_k=(1-\epsilon^{2}\lambda_k)\boldsymbol\phi_k$, so the
averaging shrinks each mode by a known factor that the consistent extension
divides back out:
\begin{equation}
\phi_k(\mathbf x_*)=\frac{1}{1-\epsilon^{2}\lambda_k}
\sum_{i\in\mathcal N_k(\mathbf x_*)}p(\mathbf x_*,i)\,\phi_k(\mathbf x_i)\ \cdot\ b(r_*).
\label{eq:nystrom}
\end{equation}

\subsection{Sparse variational inference}
\label{sec:app_svgp}

To avoid the $\mathcal O(N^3)$ cost of \eqref{eq:gp_posterior}--\eqref{eq:lml} we
use the sparse variational GP (SVGP)
\cite{titsias2009variational,hensman2013gaussian}. We introduce $M\ll N$
\emph{inducing inputs} $Z=\{\mathbf z_m\}_{m=1}^M$ with function values
$\mathbf u=f(Z)$, which are simply points of the same GP, so
$p(\mathbf u)=\mathcal N(m(Z),K_{uu})$ and $p(\mathbf f\mid\mathbf u)$ is the
usual Gaussian conditional. The inducing points follow the probability distribution 
$q(\mathbf u)=\mathcal N(\mathbf m,\mathbf S)$ ($\mathbf S=\mathbf L\mathbf
L^\top$ parameterized by its Cholesky factor) and approximate the posterior
process by $q(\mathbf f)=\int p(\mathbf f\mid\mathbf u)\,q(\mathbf u)\,d\mathbf
u$, which is Gaussian at any set of inputs $X$ with
\begin{equation}
\boldsymbol\mu_q=m(X)+K_{fu}K_{uu}^{-1}(\mathbf m-m(Z)),
\qquad
\Sigma_q=K_{ff}-K_{fu}K_{uu}^{-1}K_{uf}
+K_{fu}K_{uu}^{-1}\mathbf S\,K_{uu}^{-1}K_{uf}.
\label{eq:svgp_pred}
\end{equation}
The variational parameters $(\mathbf m,\mathbf S)$, the hyperparameters, and
optionally $Z$ itself are trained by maximizing the evidence lower bound
\begin{equation}
\mathcal L=\sum_{i=1}^{N}\mathbb E_{q(f(\mathbf x_i))}\big[\log p(y_i\mid f(\mathbf x_i))\big]
-\mathrm{KL}\big(q(\mathbf u)\,\|\,p(\mathbf u)\big)\ \le\ \log p(\mathbf y),
\label{eq:elbo}
\end{equation}
which factorizes over data points and therefore admits minibatch stochastic
optimization, reducing the cost to $\mathcal O(NM^2)$ in time and
$\mathcal O(M^2)$ in memory per step. In practice $q(\mathbf u)$ is stored in the
 parameterization $\mathbf u=L_{uu}\mathbf v$ with
$K_{uu}=L_{uu}L_{uu}^\top$, which decouples the variational parameters from the
kernel hyperparameters and conditions the optimization.

Our model uses \eqref{eq:elbo} in its latent-variable form. Each of the $d$
latent dimensions is an independent GP over the spatial coordinates (an
independent-multitask variational strategy with shared inducing locations), and
the latent vector is decoded to the $p$ observed lipid channels by an MLP, so the
GP acts as a smooth spatial \emph{prior on the decoder inputs}. Because the
decoder is nonlinear, the expected log-likelihood in \eqref{eq:elbo} has no
closed form and is estimated by a reparameterized Monte-Carlo sample
$\mathbf z\sim q(\mathbf f)$ pushed through the decoder $g_{\boldsymbol\psi}$,
giving the training objective
\begin{equation}
\mathcal L_{\text{train}}
=\underbrace{\mathbb E_{q}\big[\log p(\mathbf y\mid g_{\boldsymbol\psi}(\mathbf z))\big]}_{\text{reconstruction}}
-\ \beta\,\mathrm{KL}\big(q(\mathbf u)\|p(\mathbf u)\big)
\
\end{equation}

\subsection{Illustrative example}

To illustrate the effectiveness of the Implicit Manifold Gaussian Process when the data clearly lies on a lower-dimensional manifold, in this case a Swiss roll, we used the following function:

\begin{align}
\mathbf{x}(t,h) &= \big(r(t)\cos t,\;\; h,\;\; r(t)\sin t\big),
  \qquad r(t) = r_0 + \frac{g}{2\pi}\,t \\[4pt]
f(t) &= \sin(c\,t)
\end{align}

with $t \in [0, 2\pi T]$, $h \in [0, H]$, and $c$ = 1.5, $g$ = 0.2.

\begin{figure}[h]
    \centering
    \includegraphics[width=1.0\linewidth]{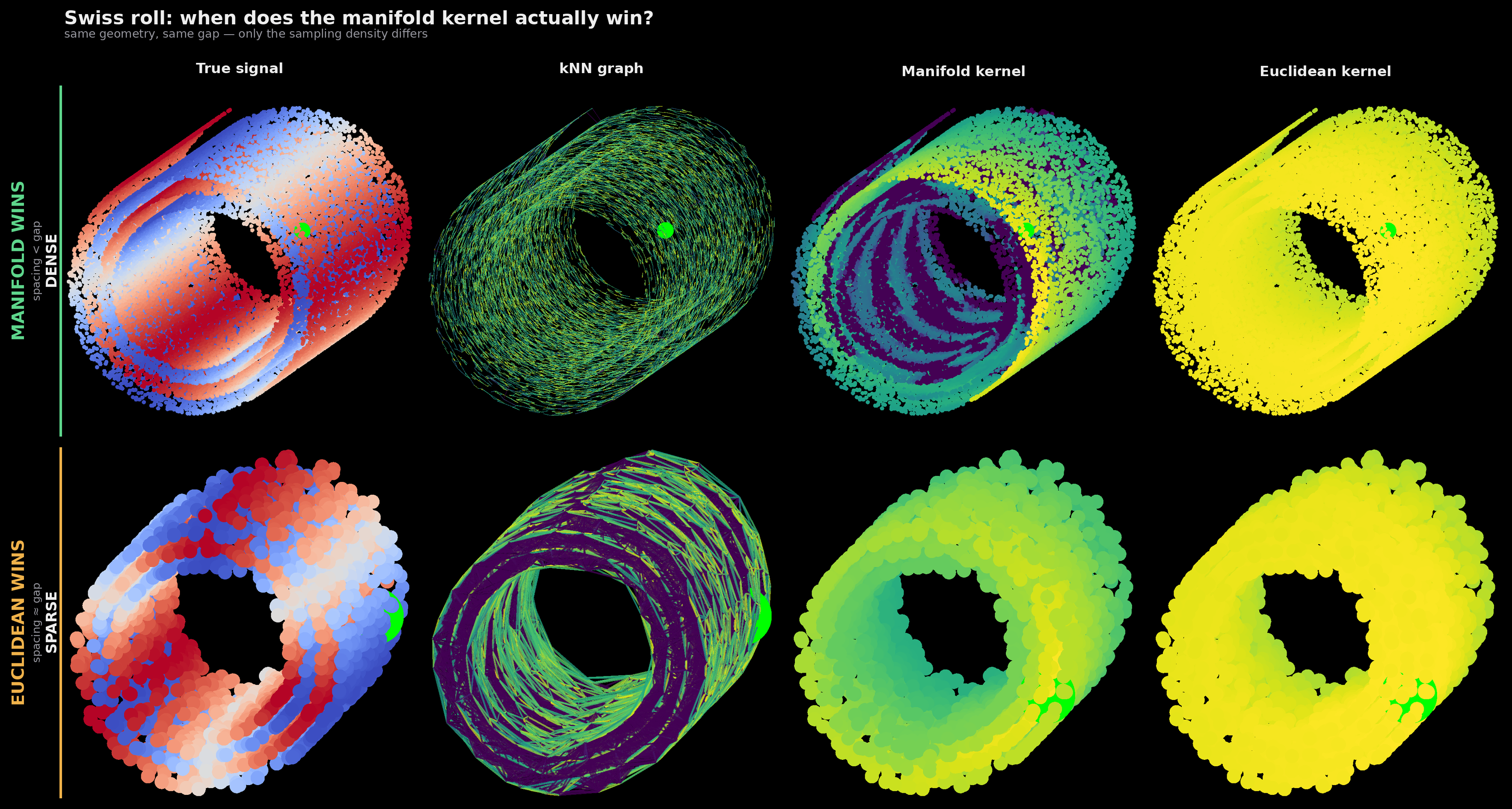}
    \caption{Illustration of the Implicit Manifold GP vs Matérn GP on the Swiss roll}
    \label{fig:swiss_roll}
\end{figure}

In Figure \ref{fig:swiss_roll}, we provide the comparison of the original signal, the constructed kNN, a heatmap of the Riemannian Manifold kernel values relative to a test point, and a heatmap of the Euclidean kernel values relative to the same test point. According to table \ref{tab:swissroll}, the Implicit Manifold GP obtains a 5.4x improvement over the Matérn kernel in terms of the Spearman correlation and Mean Square Error. Figure \ref{fig:swiss_roll} also illustrates an example where the closest points cross the implicit manifold boundary during the kNN construction and create a kernel that is an approximation of the Euclidean kernel - similar to the Allen 3D reference mouse brain image.

\begin{table}[h]
  \centering
  \begin{tabular}{llrrr}
    \toprule
    Regime & Method & RMSE & Pearson & Spearman \\
    \midrule
    \multirow{2}{*}{Dense}
      & euclidean      & 0.6735 & 0.316 & 0.286 \\
      & manifold plain & \textbf{0.1279} & \textbf{0.984} & \textbf{0.976} \\
    \midrule
    \multirow{2}{*}{Sparse}
      & euclidean      & \textbf{0.6721} & \textbf{0.259} & \textbf{0.237} \\
      & manifold plain & 0.6916 & 0.255 & 0.220 \\
    \bottomrule
  \end{tabular}
  \caption{Test-set accuracy on the swiss roll. Only the sampling density
    differs between regimes; the euclidean kernel is unchanged (0.6735 vs
    0.6721) while the manifold kernel swings by $5.4\times$.}
  \label{tab:swissroll}
\end{table}

\section{Appendix: The MALDI dataset}
\label{sec:maldi_analysis}
\begin{figure}[ht]
    \centering
    \includegraphics[width=0.7\linewidth]{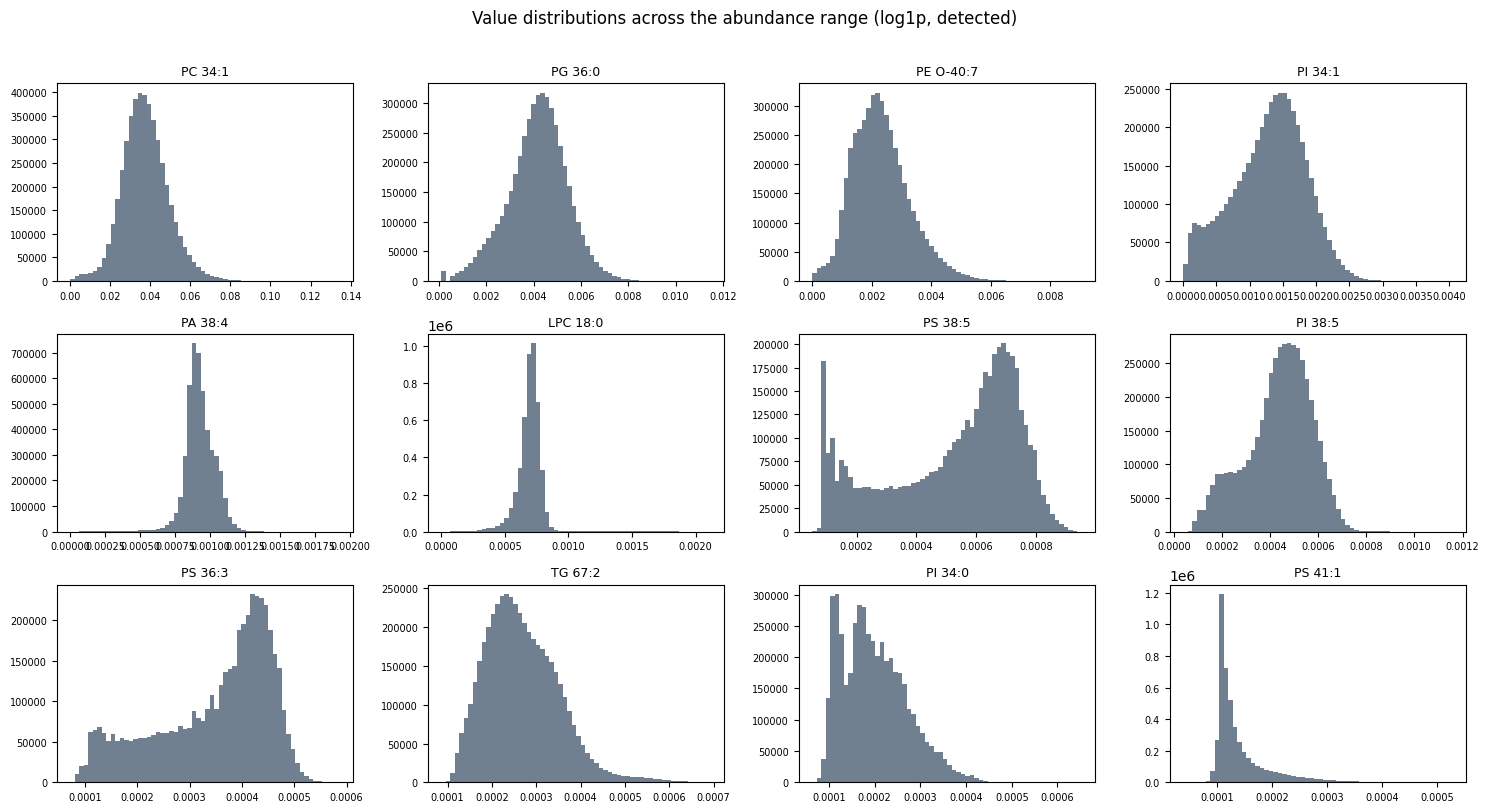}
    \caption{Value distribution across a sample of lipids}
    \label{fig:lipid_values}
\end{figure}

The MALDI dataset consists of $\sim$7 million datapoints with 173 lipid channels registered to the Allen mouse brain reference \cite{allen}, with values in the range [0, 0.14] with a median of 6e-4, representing the lipid concentration in that particular cell. Most lipids follow a roughly Gaussian distribution as seen in figure \ref{fig:lipid_values}.

\begin{figure}[ht]
    \centering
    \includegraphics[width=0.4\linewidth]{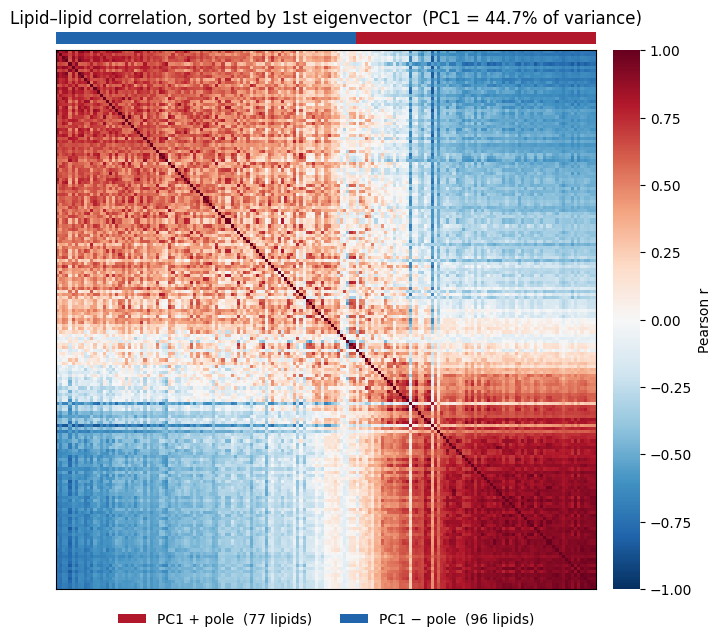}
    \caption{Lipid correlation matrix}
    \label{fig:lipid_corr}
\end{figure}

\begin{figure}[ht]
    \centering
    \includegraphics[width=1.0\linewidth]{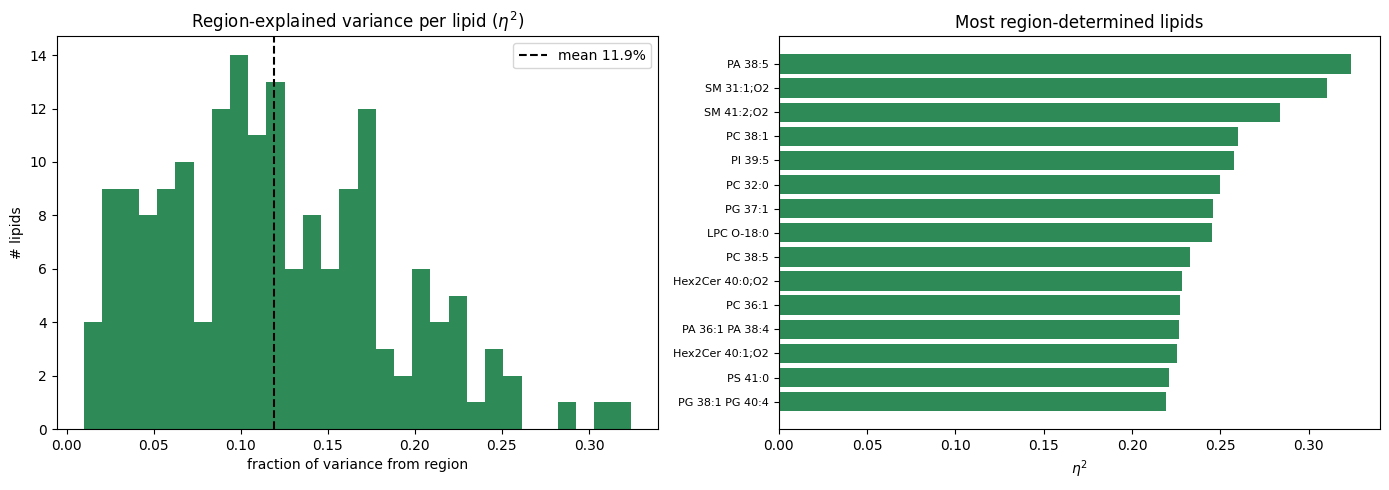}
    \caption{Lipid variance explained by the Allen anatomical region ($\eta^2$). Left: distribution across the 173 lipids (dashed line: mean, 11.9\%). Right: the 15 lipids most determined by region.}
    \label{fig:lipid_region}
\end{figure}

Two noteworthy observations about the lipid profiles are that they highly correlate with each other - in fact, sorting the correlogram in Figure \ref{fig:lipid_corr} by the first eigenvector gives us 4 clear spatial patterns which justify our choice of a 5-dimensional latent space, and that individual lipids heavily correlate with the anatomical region. Figure \ref{fig:lipid_region} shows that the median explained variance with the Atlas annotation\_id is 11.9\%, with some specific lipids (PA 38:5 going beyond 30\%).

\end{document}